\documentclass[11pt]{article}

\usepackage{acl}

\usepackage[english]{babel}
\usepackage[T1]{fontenc}
\usepackage[utf8]{inputenc}
\usepackage{times}
\usepackage{microtype}
\usepackage{inconsolata}
\usepackage{CJKutf8}

\definecolor{softgray}{HTML}{F5F5FA} 
\definecolor{darkgray}{HTML}{4A4A8A}
\usepackage{fontawesome5}
\usepackage{amsmath}
\usepackage{amssymb}
\usepackage{amsfonts}
\usepackage{mathtools}
\usepackage{bm}

\usepackage{graphicx}
\usepackage{tikz}
\usepackage{pgfplots}
\pgfplotsset{compat=1.18}
\usepackage{subcaption}
\usepackage{float}

\usepackage{booktabs}
\usepackage{tabularx}
\usepackage{multirow}
\usepackage{array}
\usepackage{makecell}
\usepackage[most]{tcolorbox}
\usepackage{fancyvrb}
\usepackage{xcolor}

\usepackage{latexsym}
\usepackage[normalem]{ulem}
\usepackage[table]{xcolor}
\usepackage{xspace}
\usepackage{url}
\usepackage{enumitem}
\usepackage{tcolorbox}
\usepackage{comment}
\usepackage{hyperref}
\usepackage{cleveref}

\definecolor{percCol}{HTML}{FFE6B3}
\definecolor{actCol}{HTML}{FFD6D6}
\definecolor{condCol}{HTML}{D6E6FF}
\definecolor{abscol}{HTML}{D6F5D6}
\definecolor{newcol}{HTML}{2E7D32}
\definecolor{boxbg}{HTML}{F5F5FA}
\definecolor{boxrule}{HTML}{4A4A8A}
\definecolor{sectiongray}{RGB}{235,235,235}
\newcommand{\cts}{\textsc{CTS}\xspace}
\newcommand{\bench}{\textsc{WorldBench}\xspace}

\newcommand{\plotorbox}[2]{%
  \IfFileExists{#1}{%
    \includegraphics[width=#2]{#1}%
  }{%
    \fbox{\parbox{#2}{\centering Missing figure file: \nolinkurl{#1}}}%
  }%
}

\title{\textsc{WorldBench}: Culturally Grounded Benchmark for Multilingual Agents}

\author{
  Leonardo Ranaldi $^{(\bullet)}$ \quad
Sherrie Shen$^{(\bullet)}$ \quad Jushi Kai$^{(\bullet,\circ)}$ \quad  Alexandra Birch$^{(\bullet)}$ \\
 	${(\bullet)}$ ILCC, School of Informatics, University of Edinburgh \\
     	${(\circ)}$ School of Artificial Intelligence, Shanghai Jiao Tong University \\
\small  \texttt{\{first\_name.last\_name\}@ed.ac.uk}
}

\begin{document}
\maketitle

\begin{abstract}
Despite the growing use of LLM-powered agents to solve multi-step tasks in complex environments, existing benchmarks rarely test state preservation, performance across languages, and application to realistic, grounded scenarios. To address these concerns, we present \bench: a comprehensive, multilingual benchmark of genuine, persona-grounded everyday workflows, where agents can act in a sandbox via structured actions. \bench comprises 1,600 tasks across seven languages and eight cultures, filtered and refined through feedback from human annotators with language- and culture-specific expertise. For evaluation, we extend metrics from previous works and introduce \textit{Constrained Task Success} (\cts), which combines natural language instructions and testbeds to score task completion, minimal modification, and other complementary metrics through deterministic and LLM-as-a-Judge evaluations. Our experiments show that frontier models reach only 49.2\% \cts, with all models demonstrating large gaps between correctness and environment preservation. We thereby show that current agents remain brittle in multilingual, agentic scenarios, especially for long-horizon tasks and under state-preservation constraints\footnote{\textbf{Code \& data:} 
\href{https://github.com/OmniAILab/WorldBench}{\faGithub\ \texttt{WorldBench}}}
\end{abstract}

\section{Introduction}

Since the introduction of LLM-powered agents to the workplace, the tasks required of them have grown increasingly complex. Compared to deterministic, one-shot tasks, agents today are expected to tackle complex problems that require interacting with environments and make use of several tools \cite{lin2025leanstarlearninginterleavethinking,boisvert2025workarenacompositionalplanningreasoningbased,wang2024officebenchbenchmarkinglanguageagents}. Completing such tasks requires identifying relevant files, interpreting contextual constraints, executing tool calls, and stopping with a valid final state \cite{wang2025odysseybenchevaluatingllmagents,yue2026statictemplatesdynamicruntime}. For instance, updating a reimbursement sheet requires locating the correct form, filling in the proper information by finding receipts scattered across emails on various dates, and submitting it to one's manager for approval.

\begin{figure*}[t]
    \centering
    \includegraphics[width=0.95\textwidth]{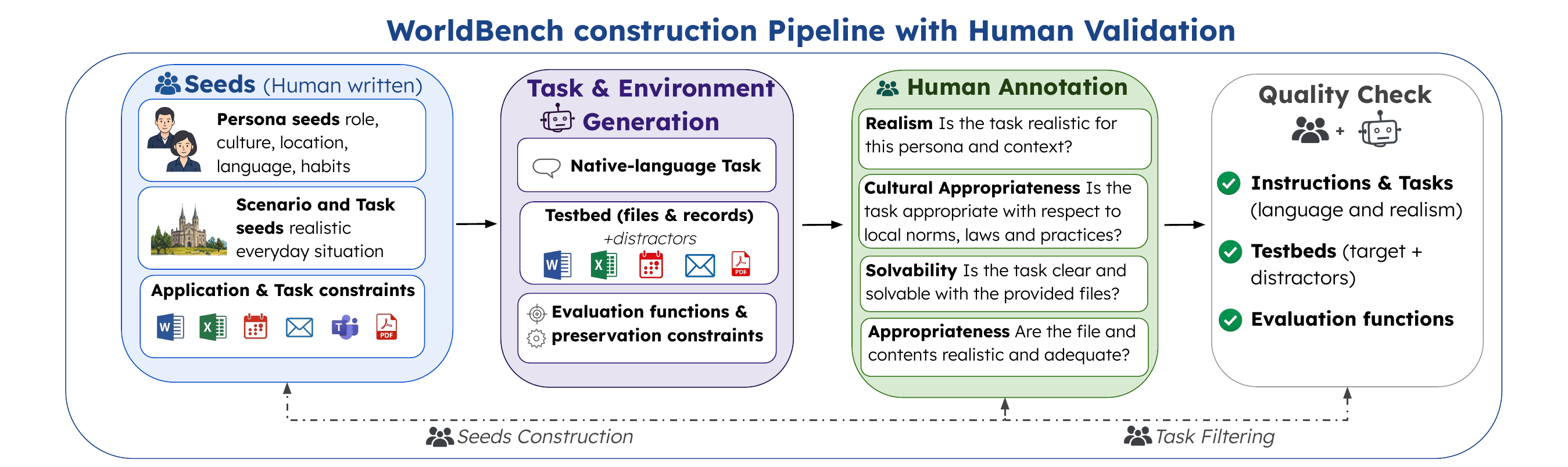}
\caption{\bench construction and validation pipeline. Human-written persona, scenario, and seeds are expanded via automated generation, then filtered and refined using language- and culture-specific annotators. }
    \label{fig:construction}
\end{figure*}

Reliable evaluation for such settings remains difficult, with agentic benchmarks focusing on web navigation \cite{deng2023mind2webgeneralistagentweb,wang-etal-2025-x}, tool use \cite{huang-etal-2024-planning-creation}, coding \cite{deng2025swebenchproaiagents}, or operating-system control \citep{zhou2024webarenarealisticwebenvironment,xie2024osworldbenchmarkingmultimodalagents,liu2025agentbenchevaluatingllmsagents}. 
While these resources have advanced the study of interactive LLMs, three problems remain underexplored. 
The first, contextual grounding: tasks are often expressed as generic instructions, whereas in real-world workflows, tasks are shaped by the role, location, language, and working habits or use cases of a specific user. An agent that ignores this context may act on the wrong assumptions.
The second, state preservation: a model may produce the requested output while accidentally changing an unrelated spreadsheet, deleting a file, or overwriting a record that should have remained untouched. Correctness-only scoring or task goal completion misses this failure, despite it being critical in practical workflows.
The third is multilingual coverage: many benchmarks are built primarily in English and then translated, leaving it an open question as to whether agents behave reliably across languages and culturally-specific scenarios.

We introduce \bench, a comprehensive and culturally grounded multilingual benchmark. Each task is grounded in a everyday scenario of a persona, then settled with a natural-language instruction, testbed, and set of evaluation functions. Agents can operate in a sandboxed environment via an action interface, observing the output, continuing for steps, and finishing either when they emit a command or when reach the iteration cap. 

Figure~\ref{fig:construction} summarises the construction and
validation pipeline. \bench combines human-generated seeds, automated generation, followed by language- and culture-specific human validation.
Unlike multilingual benchmarks derived from a shared translated task pool, each \bench setting is constructed and audited directly within its target language and cultural context. Complementing the foundational benchmarks \citet{deng2023mind2webgeneralistagentweb,xie2024osworldbenchmarkingmultimodalagents,liu2025agentbenchevaluatingllmsagents}, we introduce culturally grounded, native-language file-based workflows and evaluate both task correctness and preservation of the surrounding environment.

To improve benchmark quality, generated tasks are filtered and revised through feedback from human annotators with language- and culture-specific expertise. Annotators assess whether each task is realistic for the persona, appropriate for the local context, solvable from the supplied files, and supported by plausible testbed information. Their feedback is used to remove invalid instances and refine instructions, files, and evaluation criteria.

We design the benchmark to measure whether an agent can solve a task without damaging its workspace. Hence, we introduce \textit{Constrained Task Success} (\cts), a task-level final-state metric where a task receives \cts only when every task-specific evaluation function passes and the preservation constraint holds. The metric distinguishes nominal task completion from completion that leaves unrelated states intact. It complements intermediate trajectory metrics, which measure partial progress but do not establish whether the final environment is correct. This distinction is important in real environments containing multiple files, notes, spreadsheets, and related distractors. Accordingly, in \bench testbeds, we build testbeds that contain distractors, ambiguous records, and other constraints to identify inappropriate changes.

\bench studies multilingual robustness across seven different languages: English (US and UK), Italian, Portuguese, Spanish, French, German, and Chinese. Personas, tasks, documents, calendars, and emails are localised to each cultural and linguistic context, while the evaluation logic remains consistent. This design supports comparison across compositionally aligned settings while preserving language- and locale-specific conventions. 

We evaluate nine frontier LLM agents on \bench. The strongest model reaches 49.2\% \cts, while the weakest achieves 10.8\%. Across models, pass rate is higher than \cts, indicating that many apparently correct trajectories modify files outside the target. We also find a stable language gradient, with English settings leading and Chinese trailing across every model. Finally, we conduct a failure analysis which shows that wrong final states, edits, iteration-limit hits, malformed actions, and execution errors contribute to the limitations.
The contributions of this work are:
\begin{itemize}[itemsep=0.1pt,topsep=2.5pt,leftmargin=0.2cm]%
\item We present \bench, a multilingual agentic benchmark for culturally grounded tasks, with instances refined by human annotators across seven languages and eight cultures.

\item We define a task construction pipeline that turns personas, applications, and constraint seeds into localised instructions, concrete testbeds, and executable evaluators.

\item We introduce \cts, a headline metric that pairs task correctness with preservation of non-target files, together with diagnostic metrics that expose collateral edits, long trajectories, malformed actions, and failed executions.

\item We provide an empirical evaluation of nine LLM agents and show that they remain brittle under long-horizon execution, multilingual localisation, and preservation constraints.
\end{itemize}

\section{Related Work}

\paragraph{Language agent benchmarks.}
Recent work evaluates LLM agents in interactive environments that require planning, tool use, and feedback-driven decision-making. Representative settings include web navigation \citep{deng2023mind2webgeneralistagentweb,zhou2024webarenarealisticwebenvironment,koh-etal-2024-visualwebarena,wang-etal-2025-x}, shopping and task-oriented interaction \citep{yao2023webshopscalablerealworldweb,kim2026agenticshopbenchmarkingagenticproduct}, operating-system control \citep{xie2024osworldbenchmarkingmultimodalagents}, multi-turn tool use \citep{ma2024agentboardanalyticalevaluationboard,liu2025agentbenchevaluatingllmsagents}, and executable programming environments \citep{yang2023intercodestandardizingbenchmarkinginteractive,deng2025swebenchproaiagents}. \bench provides a complementary setting in which agents must act for a culturally-situated persona within a file-based world while preserving files outside the target set.
Recent file-based and multi-application benchmarks move agent evaluation towards long-horizon workflows with heterogeneous tools and state \citep{wang2025odysseybenchevaluatingllmagents}.

\paragraph{Document understanding and file-based workflows.}
Document understanding benchmarks traditionally focus on extracting information from forms, receipts, invoices, tables, or visually rich documents \citep{mathew2021docvqadatasetvqadocument,gong2025mhierragmultimodalragvisualrich,zhang2026parsebenchdocumentparsingbenchmark}. These tasks test document interpretation but do not require agents to maintain a workspace, modify artefacts, or preserve unrelated state during multi-step execution. Recent agent benchmarks address richer file-based workflows, but are typically centred on English scenarios and do not make culturally grounded multilingual task construction the core design goal. \bench targets this gap by combining documents, spreadsheets, messages, calendar records, PDFs, shell commands, and generated notes into executable tasks that are built and audited across languages.

\paragraph{Synthetic benchmark construction.}
Synthetic data generation has been widely used to create instruction-following corpora and task variants from compact seeds \citep{wang2023selfinstructaligninglanguagemodels,xu2025wizardlmempoweringlargepretrained,chim-etal-2025-evaluating,gill-etal-2025-lost}. For agent evaluation, generation is more demanding because each task must contain a coherent state, a solvable instruction, and an executable evaluator \cite{fang2025comprehensivesurveyselfevolvingai,10.1007/978-3-031-93418-6_9}. \bench treats generated tasks as complete artefacts: each instance includes localised text, files, and evaluation functions. The benchmark is then filtered and refined via feedback from human annotators, who assess task authenticity, cultural reliability, file sufficiency, and evaluator correctness.

\paragraph{Multilingual agent evaluation.}
LLM evaluation is still English-centred. Recent multilingual agent benchmarks show that performance and safety can degrade when agents are evaluated beyond English \citep{wang-etal-2025-x}. However, multilingual coverage is often obtained by translating existing English-centred tasks into other languages \citep{hofman-etal-2026-maps}. This design is useful for controlled comparison, but it can leave the underlying scenarios, user assumptions, and artefacts tied to the source language. \bench follows a different construction principle: tasks are generated, localised, and human-annotated from the start within each language setting. Personas, tasks, files, and evaluation criteria are therefore aligned with the target language and cultural context, while the evaluation logic remains shared across languages.

\section{\bench Benchmark}
\label{sec:benchmark}

Agentic evaluation must account for the worlds in which actions take effect. While recent agents can follow instructions in isolated tool-use settings, realistic workflows require coordinating operations across heterogeneous environments, where different cultures, interactions, and languages may be present, all while preserving the surrounding state. This is challenging because an agent must identify the relevant artefacts, operate within a constrained space, and terminate with a valid final state.

\begin{figure*}[t]
    \centering
    \includegraphics[width=0.95\textwidth]{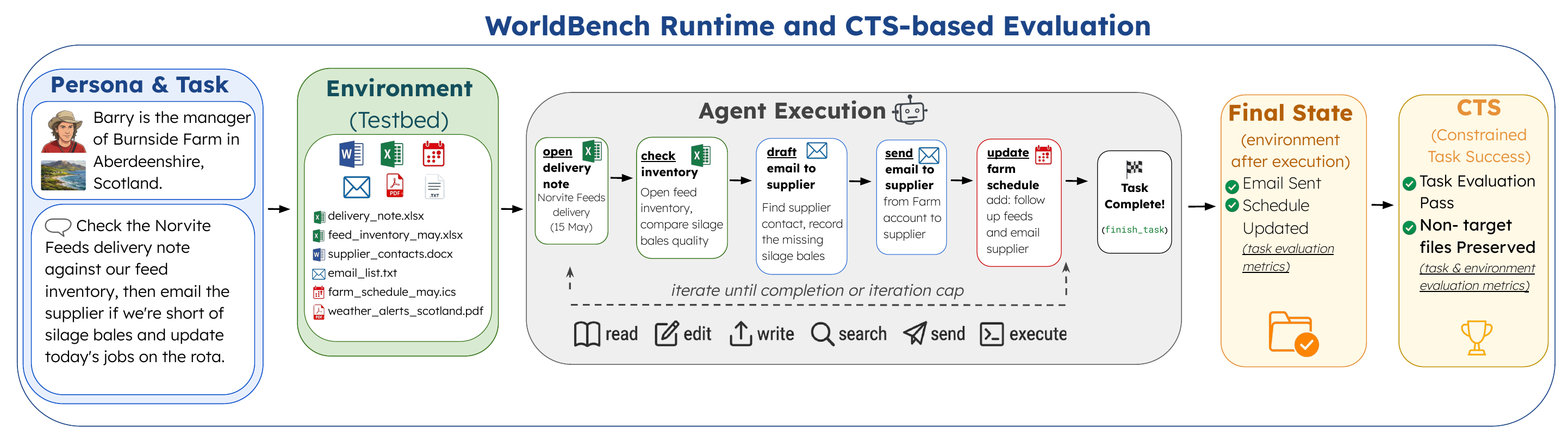}
    \caption{\bench runtime and final-state evaluation. The agent executes structured actions, after which the resulting state is evaluated for task completion and preservation of non-target files.}
    \label{fig:runtime_fig}
\end{figure*}

To evaluate these capabilities, we introduce \bench, a culturally-grounded multilingual benchmark for agentic tasks. \bench contains 1,600 tasks across seven languages and eight cultures. Each task pairs a persona-grounded instruction with a sandbox, structured action interface, and final-state evaluation functions. The benchmark is constructed from human-written seeds, materialised into task environments, and assessed through a stratified human audit with language- and culture-specific annotators. Figure~\ref{fig:construction}
summarises the construction and Figure~\ref{fig:runtime_fig} execution pipelines. We describe the task formulation
(\S~\ref{sec:task-formulation}), persona and language design
(\S~\ref{sec:personas-languages}), tool environment
(\S~\ref{sec:tool-environment}), and autonomous workflow
(\S~\ref{sec:autonomous-workflow}). We then present benchmark construction
(\S~\ref{sec:construction}) and the evaluation protocol
(\S~\ref{sec:evaluation-protocol}).  

\subsection{Task Formulation}
\label{sec:task-formulation}

We formulate an agentic task as a goal-directed interaction between an LLM-based agent and a sandboxed environment. Each task consists of a target-language instruction associated with a persona situated in an environment with a state, an available tool set, and a set of final-state evaluation criteria. These elements correspond to the persona \& task, environment, agent-execution, and final-state components in
Figure~\ref{fig:runtime_fig}. The persona identifies the user on whose behalf the agent acts and grounds the task in a role, location, and working context. The task instruction specifies the goal to be completed, while the initial state contains the relevant artefacts and related distractors. The evaluation define the properties that must hold when execution ends. This formulation is designed to test grounded execution since the agent cannot solve a task by producing an answer in isolation; instead, it must inspect the available artefacts, select appropriate tools, execute actions, and leave the environment in a valid final state. This makes \bench suitable for evaluating long-horizon behaviour and tool coordination. 

\subsection{Personas and Native-Language Task Construction}
\label{sec:personas-languages}

A central goal of \bench is to evaluate agents in culturally grounded settings. \bench covers persona-related tasks in seven different languages: English, Italian, Portuguese, Spanish, French, German, and Chinese. The benchmark is organised by language and culturally grounded via the persona's location and context.

Each persona specifies a concrete user, including name, role, locale, language, and relevant context. These attributes determine which activities, artefacts, registers, dates, currencies, organisations, and assumptions are appropriate for the task. 

The personas and tasks are not obtained by translating a single language pool. They are constructed within the target setting; hence, names, currencies, file contents, documents, and contextual assumptions are aligned with the corresponding locale. The evaluation logic is shared across settings, enabling comparison while preserving language- and location-specific variation.

\subsection{Tool Environment \& Workflow}
\label{sec:tool-environment}

\bench exposes a heterogeneous tool environment that covers the main artefact types required by the tasks. Each tool family is associated with a restricted set of operations, Table~\ref{tab:tools} summarises the tool families used in the benchmark.

\begin{table}[h]
\centering
\small
\setlength{\tabcolsep}{3pt}
\begin{tabularx}{\columnwidth}{@{}lX@{}}
\toprule
\textbf{Tool family} & \textbf{Operations} \\
\midrule
Spreadsheet & Read tabular records and edit cell values. \\
Document & Read, create, and update text documents. \\
PDF & Read PDF content and support conversion-oriented operations. \\
Messaging & Inspect message records and compose messages. \\
Calendar & List events and create non-overlapping entries. \\
Shell & Execute filesystem commands inside the sandbox. \\
System & Manage global control actions and terminate execution. \\

\bottomrule
\end{tabularx}
\caption{Tools and operations in \bench. }
\label{tab:tools}
\end{table}

The tool environment creates a broad action space, but only a subset of operations is appropriate at any point in a trajectory. The agent must therefore decide which artefact to inspect, which tool to use, and how to parameterise the next action. To trace execution, we record errors in tool choice, invalid arguments, repeated operations, and premature termination.

\subsection{Autonomous Workflow}
\label{sec:autonomous-workflow}

We model the interaction as a transition system
$\mathcal{W} = (S, A, O, \delta)$, where $S$ is the space of environment states, $A$ is the structured action space, $O$ is the observation space, and $\delta:S \times A \rightarrow S \times O$ is the transition function. At step $t$, the agent selects an action $a_t \in A$. The environment executes the action, updates the sandbox state, and returns an observation $o_t$. The execution history is represented as $H_t = [(a_1,o_1),\ldots,(a_{t-1},o_{t-1})]$.
Figure~\ref{fig:runtime_fig} shows this execution protocol. The agent receives a persona-grounded task, interacts with the testbed via structured actions, and is evaluated from the resulting final state.

The agent conditions its next action on the task instruction, available tool descriptions, and $H_t$. Observations are rendered in text and preserve the information needed for subsequent decisions. Depending on the operation, they may contain document content, spreadsheet values, command outputs, calendar entries, generated messages, or execution errors. Execution proceeds until the agent emits a termination action or reaches the iteration cap. The final state is then passed to the evaluation functions. This workflow allows \bench to measure whether the agent can build a coherent action trajectory, use observations effectively, and stop after reaching the intended final state.

\subsection{Benchmark Construction}
\label{sec:construction}

\bench is constructed through a seed-driven construction pipeline comprising
task synthesis, testbed synthesis, and human audit. The purpose is to create
executable tasks that are heterogeneous, culturally grounded, and auditable.

\paragraph{Task Synthesis}

Construction begins from three types of human-written seeds. A \textit{persona seed} specifies the user's role, location, language, and working context. A \textit{scenario seed} describes a plausible activity for that persona. An \textit{application-and-constraint seed} identifies the required tools, intended outputs, forbidden outcomes, and preservation
requirements.

These elements are expanded into a native-language instruction and a specification of the artefacts required by the task. The resulting scenario must be reliable for the persona and solvable through the supported tool environment.

\paragraph{Testbed Synthesis}

The testbed synthesis stage materialises the task environment. Spreadsheets, PDFs, documents, calendars and text files are populated with task-relevant values; documents contain persona-consistent prose; mailboxes contain message records; and calendars contain events required by scheduling tasks. Each testbed also includes 20 distractor artefacts containing similar filenames, overlapping entities, related records, or outdated versions of target documents. Distractors increase the need for precise artefact selection and support the evaluation of workspace preservation.

\paragraph{Human Audit}
\label{sec:human}

The candidate task pool is reviewed by annotators with language- and culture-specific expertise before final inclusion. Annotators examine whether each task is realistic for the persona, appropriate for its local context, and solvable from the supplied artefacts. For each file, they also assess whether the type and amount of information are plausible for that document format.

The audit identifies problems that automatic validation cannot reliably capture, including culturally implausible scenarios, unnatural document contents, insufficient evidence, and evaluation criteria that under-specify the intended outcome. Appendix~\ref{app:audit} reports the construction statistics, review procedure, and annotation questions.

\subsection{Evaluation Protocol}
\label{sec:evaluation-protocol}

Each task is evaluated from the final state. \bench combines deterministic evaluation with LLM-as-a-judge evaluation. Deterministic functions check properties that can be verified directly, including required or forbidden content, spreadsheet cell values, file existence, and calendar consistency. Judge-based functions are used for open-ended artefacts, such as messages or notes.

A task passes when all its task-specific evaluation functions succeed. These functions may inspect several properties across one or more files,
spreadsheet cells, calendar entries, or message artefacts. Pass rate therefore
measures complete satisfaction of the task-specific final-state criteria.

Pass rate does not establish whether the surrounding workspace remains unchanged. \bench individually checks whether any non-target file differs between the initial and final sandbox states. Preservation rate measures the proportion of tasks for which all pre-existing non-target files remain intact. Their
conjunction yields \cts, which we formally define in \S~\ref{sec:metrics}.

\paragraph{Evaluation Functions}
\label{sec:evalfuncs}

Each task is connected to one or more evaluation functions, either deterministic or LLM-based. The deterministic functions verify exact properties of the final filesystem, while the judge-based functions evaluate open-ended messages and notes against a task-specific rubric. 
Table~\ref{tab:evalfuncs} summarises the evaluation functions; implementation details in
Appendices~\ref{app:judge} and \ref{app:det_eval}.

\begin{table}[h]
\centering
\small
\setlength{\tabcolsep}{3pt}
\begin{tabularx}{\columnwidth}{@{}lX@{}}
\toprule
\textbf{Function} & \textbf{What it checks} \\
\midrule
\texttt{contain} & Required text appears in a target file. \\
\texttt{not\_contain} & Prohibited text is absent from a target file. \\
\texttt{excel\_cell\_value} & A spreadsheet cell contains the expected value. \\
\texttt{file\_exist} & A required output file has been created. \\
\texttt{calendar\_no\_overlap} & Calendar events satisfy a non-overlap constraint. \\
\texttt{evaluate\_email} & A generated email satisfies the rubric. \\
\texttt{evaluate\_note} & A generated note satisfies the rubric. \\
\bottomrule
\end{tabularx}
\caption{Evaluation functions used in \bench.}
\label{tab:evalfuncs}
\end{table}

\begin{table*}[h]
\centering
\small
\setlength{\tabcolsep}{4pt}
\begin{tabular}{@{}lccccccc@{}}
\toprule
\textbf{Model} &
\textsc{CTS}$\uparrow$ &
\textsc{Pass}$\uparrow$ &
\textsc{Presv.}$\uparrow$ &
\textsc{Steps}$\downarrow$ &
\textsc{Malf./task}$\downarrow$ &
\textsc{HitMax}$\downarrow$ &
\textsc{Clean}$\uparrow$ \\
\midrule
Gemini-3.1-Pro &
\textbf{49.2} & 59.3 & \textbf{82.5} & 8.7 & \textbf{0.40} & \textbf{10\%} & \textbf{88\%} \\
GPT-5 & 48.8 & \textbf{60.0} & 81.3 & \textbf{7.5} & 0.60 & 17\% & 82\% \\
Qwen-3-32B & 48.0 & 58.5 & 80.5 & 8.2 & 0.75 & 19\% & 78\% \\
GPT-4o & 38.6 & 51.7 & 73.9 & 10.4 & 0.90 & 17\% & 79\% \\
Gemini-3.5-Flash & 34.6 & 50.3 & 66.0 & 12.6 & 1.70 & 25\% & 68\% \\
Llama-3.3-70B & 24.1 & 40.9 & 58.7 & 13.9 & 2.60 & 34\% & 58\% \\
Llama-3.1-8B & 12.5 & 25.2 & 49.6 & 18.2 & 4.50 & 52\% & 41\% \\
Qwen-3-4B & 11.6 & 23.5 & 48.0 & 18.8 & 4.80 & 55\% & 38\% \\
EuroLLM-9B & 10.8 & 21.5 & 50.2 & 19.5 & 5.10 & 58\% & 35\% \\
\bottomrule
\end{tabular}
\caption{Macro-averaged results on \bench across eight cultures and all
tool sets. \textsc{CTS} combines complete task-level evaluator success
with preservation of the non-target workspace; \textsc{Pass} and
\textsc{Presv.} report these components separately. \textsc{Steps} is
the mean number of actions on passed tasks. \textsc{Malf./task},
\textsc{HitMax}, and \textsc{Clean} denote malformed actions per task,
iteration-cap hits, and termination via \texttt{finish\_task}.}
\label{tab:headline}
\end{table*}

\section{Experiments}
\label{sec:experimental-setup}

\subsection{Metrics}
\label{sec:metrics}

The headline metric is Constrained Task Success (\cts). For each task $t$, $\mathrm{pass}(t)$ is true when all task-specific evaluation functions succeed on the final sandbox state computed once for the complete task. For instance, a task requiring both a spreadsheet update and an email passes only if both outputs satisfy their respective evaluators.

We define $\mathrm{preserve}(t)$ as the condition in which every non-target file remains unchanged between the initial and final states. Target files are identified from the task evaluators; all other pre-existing files must remain present and byte-identical. Then,

$$
\cts(t) =
\mathrm{pass}(t) \wedge \mathrm{preserve}(t).
$$

\cts extends final-state task success
\citep{wang2024officebenchbenchmarkinglanguageagents,
xie2024osworldbenchmarkingmultimodalagents}
with a workspace-preservation condition. The check is binary and operates at file level: any modification to a non-target file causes failure, while changes within target files are assessed by task-specific evaluators.

We report pass rate, preservation rate, mean steps on solved tasks, malformed actions per task, execution failures, iteration-cap hits, and clean termination rate. The gap between pass rate and \cts captures tasks that satisfy the final-state criteria but violate preservation. We further report \cts by language and tool family, pass rate by evaluation function, and trajectory-level events linking collateral edits to specific actions.

\subsection{Models and Protocols}

We evaluate nine LLM-based agents: Gemini-3.1-Pro, Gemini-3.5-Flash, GPT-5, GPT-4o, Qwen-3-32B, Qwen-3-4B, Llama-3.3-70B, Llama-3.1-8B, and EuroLLM-9B. All agents use the same prompts, action schema, environment, and configurations. Details are in Appendix \ref{app:model_versions}.

Each task is run once per model, using the protocol in Appendix \ref{app:implementation}. We record the trajectory, final environment state, evaluation outcomes, and diagnostic events, reporting percentages for \cts, pass rate, preservation, iteration-cap hits, and clean terminations, as well as mean counts for malformed actions and mean trajectory on solved tasks.

\section{Results}
\label{sec:results}

\begin{table*}[h]
\centering
\small
\setlength{\tabcolsep}{4pt}
\begin{tabular}{@{}lccccccccc@{}}
\toprule
\textbf{Model} &
\textsc{EN-US} &
\textsc{EN-UK} &
\textsc{IT} &
\textsc{PT} &
\textsc{ES} &
\textsc{FR} &
\textsc{DE} &
\textsc{ZH} &
\textsc{Avg.} \\
\midrule
Gemini-3.1-Pro &
\textbf{58.4} & 56.8 & \textbf{49.1} & \textbf{46.9} & \textbf{49.9} & \textbf{47.7} & 46.2 & 38.6 & \textbf{49.2} \\
GPT-5 & 57.4 & \textbf{56.9} & 48.1 & 45.7 &
48.9 & 46.9 & \textbf{47.7} & 38.8 & 48.8 \\
Qwen-3-32B & 56.0 & 55.5 & 46.5 & 44.5 & 47.0 & 45.5 & 45.5 & \textbf{43.5} & 48.0 \\
GPT-4o & 44.7 & 43.4 & 40.4 & 36.8 & 39.7 & 39.7 & 35.4 & 28.7 & 38.6 \\
Gemini-2.5-Pro & 43.5 & 41.8 & 38.0 & 36.7 & 38.3 & 35.8 & 34.7 & 28.6 & 37.4 \\
Gemini-3.5-Flash & 38.7 & 38.4 & 31.7 & 31.5 & 32.6 & 33.6 & 30.2 & 25.7 & 34.6 \\
Llama-3.3-70B &
28.6 & 27.6 & 22.5 & 23.5 &
26.3 & 21.8 & 23.5 & 18.0 &
24.1 \\
Llama-3.1-8B & 15.5 & 15.0 & 12.0 & 11.5 & 12.5 & 11.8 & 12.2 & 9.5 & 12.5 \\
Qwen-3-4B & 13.8 & 13.3 & 10.0 & 11.1 &
10.7 & 9.8 & 10.3 & 13.8 & 11.6 \\
EuroLLM-9B & 12.5 & 12.0 & 11.2 & 10.5 & 11.0 & 10.8 & 10.9 & 7.5 & 10.8 \\
\bottomrule
\end{tabular}
\caption{Per-language \cts scores.}
\label{tab:per_language}
\end{table*}

\subsection{Main Results}

Gemini-3.1-Pro, GPT-5, and Qwen-3-32B lead the pack at 49.2\%, 48.8\%, and 48.0\% \cts respectively. GPT-4o and Gemini-3.5-Flash follow at 38.6\% and 34.6\% \cts, with the remaining models scoring between 24.1\% and 10.8\% \cts.

\begin{figure}[h]
    \centering
\plotorbox{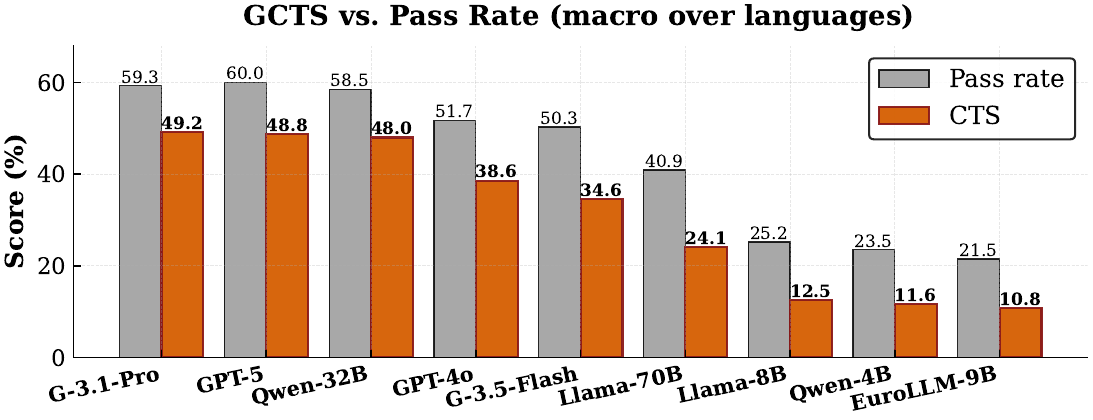}{\columnwidth}
    \caption{\cts vs pass rate. The arrows represent the penalty introduced by the preservation constraint.}
    \label{fig:headline}
\end{figure}

Table~\ref{tab:headline} and Figure~\ref{fig:headline} show that current LLM-based agents struggle on \bench. Gemini-3.1-Pro, GPT-5, and Qwen-3-32B achieve the highest scores, but their \cts still remain under 50\%. GPT-4o, Gemini-3.5-Flash and Llama-3.3-70B rank in the middle, and the rest of the models perform worse. These results indicate that strong capability does not directly translate into reliable execution once tasks are grounded in a context-dependent world.

\paragraph{The Preservation Gap}
Table \ref{tab:headline} displays that for every model, pass rate exceeds \cts by a broad margin. The gap ranges from 10.1 for Gemini-3.1-Pro to 16.8 for Llama-3.3-70B and increases as overall capability decreases. In less performant models, the test pass rate is low enough that few trajectories survive long enough to make a change. The difference between pass rate and \cts embodies the lack of environmental preservation: agents that complete the process while leaving changes in files that should have remained unedited.

\begin{figure}[h]
    \centering
    \plotorbox{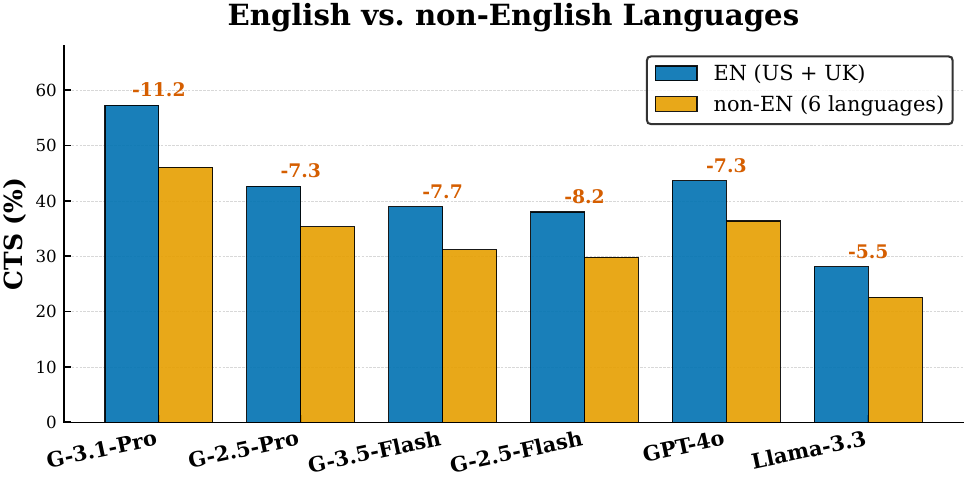}{0.9\columnwidth}
    \caption{English versus non-English \cts per model. 
    }
    \label{fig:engap}
\end{figure}

\subsection{Multilingual Robustness}

Table~\ref{tab:per_language} and Figure~\ref{fig:engap} show that English generally yield the highest \cts, while Chinese is among the most difficult settings for most models.
The Qwen models are the main exception: their decline on Chinese is smaller, and Qwen-3-32B gets the highest Chinese. 
These differences may reflect multilingual instruction following, localised document conventions, and variation in difficulty across constructed settings. We interpret them as setting-level performance gaps. In Appendix~\ref{app:cross-locale} we discuss the limits of cross-locale comparability.

\begin{figure}[h]
    \centering
    \plotorbox{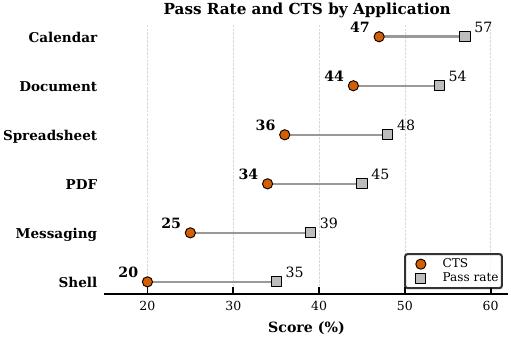}{0.85\columnwidth}
    \caption{Pass rate and \cts by application family; connecting segments indicate preservation gaps.}
    \label{fig:perapp}
\end{figure}

\subsection{Application and Evaluator Effects}

Figure~\ref{fig:perapp} compares pass rate and \cts across applications. Calendar and document tasks achieve higher scores, while messaging and shell tasks show lower \cts and larger preservation gaps. Their difficulty therefore reflects both incomplete execution and collateral modifications.

\begin{figure}[h]
    \centering
\plotorbox{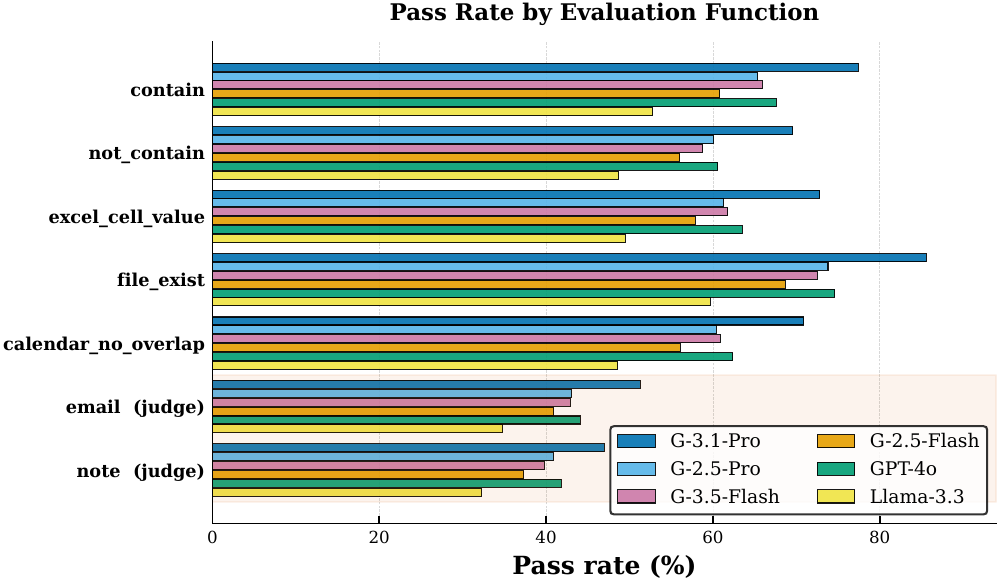}{\columnwidth}
    \caption{Pass rate by evaluation function.}
    \label{fig:pereval}
\end{figure}

Figure~\ref{fig:pereval} shows that deterministic functions, including file-existence and spreadsheet-cell checks, achieve higher pass rates than the email and note evaluators. Open-ended outputs therefore remain more difficult to satisfy. Appendices~\ref{app:judge} and \ref{app:judge-robustness} report the judging protocol and robustness analysis.

\begin{figure}[h]
    \centering
    \plotorbox{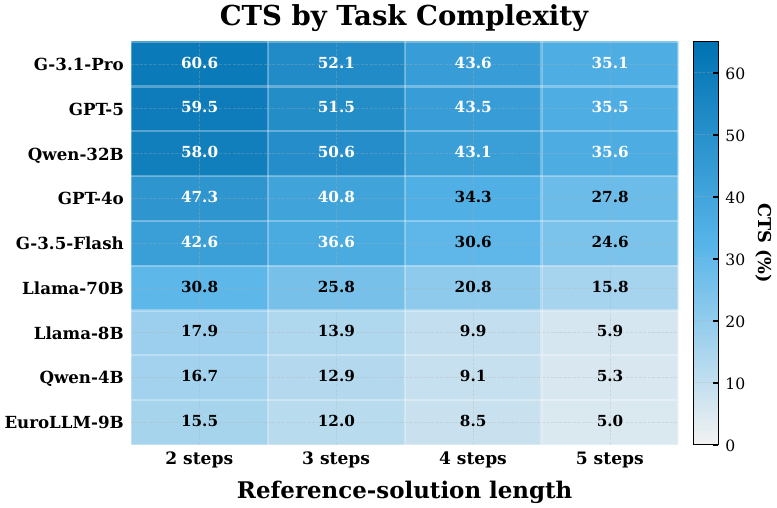}{\columnwidth}
    \caption{\cts by model and reference-solution-length.}
    \label{fig:complexity-cts}
\end{figure}

\paragraph{Task complexity.}
Figure~\ref{fig:complexity-cts} reports \cts for each model across the four reference-solution-length bins. Performance generally declines as tasks require more substantive actions. This analysis measures intrinsic task complexity, while Appendix~\ref{app:trajectory-length} examines the trajectories produced by agents.

\subsection{Failure Modes}

\begin{figure}[h]
    \centering
    \plotorbox{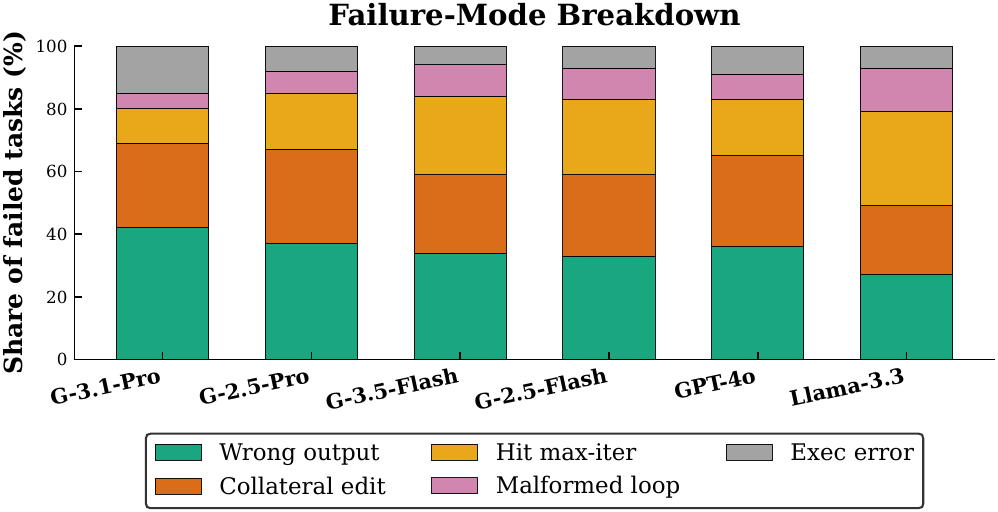}{\columnwidth}
    \caption{Failure-mode breakdown per task per model.}
    \label{fig:failures}
\end{figure}

To examine how agents fail, we classify each failed trajectory into one of five categories: wrong output, collateral edit, iteration-cap hit, malformed loop, or execution error. Each trajectory is assigned one failure mode according to the final outcome and recorded execution. Figure~\ref{fig:failures} reports the share of failed tasks assigned to each category for every model, showing that wrong output is the largest category for stronger models, with collateral edits remaining substantial across all agents. Iteration-cap hits increase as performance decreases and form the largest category for the smallest models, accounting for up to 43\% of failures in the case of EuroLLM-9B. Table~\ref{tab:headline} shows that lower-\cts models execute more steps, produce more malformed actions, terminate cleanly less often, and reach the iteration cap continually (details in Appendix~\ref{app:step-budget}).

Appendix~\ref{app:error-patterns} provides a manual error analysis and Appendix~\ref{app:trajectory-length} reports trajectory-length results for all models. Finally, Appendix~\ref{app:examples} reports some execution examples.

\section{Discussion}
\label{sec:discussion}

The results point to three limitations of current agents. First, agents lack a reliable notion of workspace safety. While they can identify a target artefact, they still add noise in neighbouring files. Second, long-horizon execution remains brittle. Once a trajectory extends beyond the solution path, errors compound quickly. Third, multilingual performance remains unstable. In parallel, these findings suggest that agents need stronger state tracking, explicit preservation objectives, and better localisation-aware action grounding. The nature of \bench is also important because it allows for new personas, languages, and patterns to be added through seeds, and it also allows fresh tasks to be created when contamination is suspected. 
\section{Conclusion}

We introduced \bench, a multilingual, persona-grounded benchmark for executable agentic tasks. \bench materialises testbeds, evaluates the final system state using deterministic and judge-based functions, and uses \cts to enforce both correctness and preservation. We experiment with different LLM-based agents; the best model does not surpass 50\% \cts. All models exhibit a notable preservation gap, and performance degrades outside English. These results show that environment preservation and multilingual robustness are central requirements for reliable agents and that both should be measured explicitly.

\section*{Limitations}

\bench evaluates structured, file-based workflows via a fixed action interface. This supports reproducible experiments on documents, spreadsheets, calendars, and filesystem operations, but excludes visual desktop control and live web interaction, which have not been included in this version to maximise reproducibility, but are intended to be incorporated in future development versions.

Two open-ended evaluators use an LLM judge, whose agreement with human annotations and alternative judges is examined in Appendix~\ref{app:judge-robustness}. Each task is executed once per model under a fixed decoding configuration.

\section*{Ethics Statement}

\bench uses synthetic personas and synthetic data. The benchmark is designed to avoid private information and personally identifying records from real users. The main ethical risks concern the evaluation of systems that may be deployed in sensitive workplace contexts. We emphasise preservation, auditability, and failure reporting. During manuscript preparation, an LLM was used for drafting and editing support. The authors reviewed all text and remain fully responsible for the scientific claims, results, and final content.

\bibliography{custom}

\begin{thebibliography}{27}
\providecommand{\natexlab}[1]{#1}

\bibitem[{Alismail and Lanquillon(2025)}]{10.1007/978-3-031-93418-6_9}
Ahmad Alismail and Carsten Lanquillon. 2025.
\newblock A survey of llm-based methods for synthetic data generation and the rise of agentic workflows.
\newblock In \emph{Artificial Intelligence in HCI}, pages 119--135, Cham. Springer Nature Switzerland.

\bibitem[{Boisvert et~al.(2025)Boisvert, Thakkar, Gasse, Caccia, Chezelles, Cappart, Chapados, Lacoste, and Drouin}]{boisvert2025workarenacompositionalplanningreasoningbased}
Léo Boisvert, Megh Thakkar, Maxime Gasse, Massimo Caccia, Thibault Le Sellier~De Chezelles, Quentin Cappart, Nicolas Chapados, Alexandre Lacoste, and Alexandre Drouin. 2025.
\newblock \href {https://arxiv.org/abs/2407.05291} {Workarena++: Towards compositional planning and reasoning-based common knowledge work tasks}.
\newblock \emph{Preprint}, arXiv:2407.05291.

\bibitem[{Chim et~al.(2025)Chim, Ive, and Liakata}]{chim-etal-2025-evaluating}
Jenny Chim, Julia Ive, and Maria Liakata. 2025.
\newblock \href {https://doi.org/10.1162/coli_a_00540} {Evaluating synthetic data generation from user generated text}.
\newblock \emph{Computational Linguistics}, 51(1):191--233.

\bibitem[{Deng et~al.(2025)Deng, Da, Pan, He, Ide, Garg, Lauffer, Park, Pasari, Rane, Sampath, Krishnan, Kundurthy, Hendryx, Wang, Bharadwaj, Holm, Aluri, Zhang, Jacobson, Liu, and Kenstler}]{deng2025swebenchproaiagents}
Xiang Deng, Jeff Da, Edwin Pan, Yannis~Yiming He, Charles Ide, Kanak Garg, Niklas Lauffer, Andrew Park, Nitin Pasari, Chetan Rane, Karmini Sampath, Maya Krishnan, Srivatsa Kundurthy, Sean Hendryx, Zifan Wang, Vijay Bharadwaj, Jeff Holm, Raja Aluri, Chen Bo~Calvin Zhang, and 3 others. 2025.
\newblock \href {https://arxiv.org/abs/2509.16941} {Swe-bench pro: Can ai agents solve long-horizon software engineering tasks?}
\newblock \emph{Preprint}, arXiv:2509.16941.

\bibitem[{Deng et~al.(2023)Deng, Gu, Zheng, Chen, Stevens, Wang, Sun, and Su}]{deng2023mind2webgeneralistagentweb}
Xiang Deng, Yu~Gu, Boyuan Zheng, Shijie Chen, Samuel Stevens, Boshi Wang, Huan Sun, and Yu~Su. 2023.
\newblock \href {https://arxiv.org/abs/2306.06070} {Mind2web: Towards a generalist agent for the web}.
\newblock \emph{Preprint}, arXiv:2306.06070.

\bibitem[{Fang et~al.(2025)Fang, Peng, Zhang, Wang, Yi, Zhang, Xu, Wu, Liu, Li, Ren, Aletras, Wang, Zhou, and Meng}]{fang2025comprehensivesurveyselfevolvingai}
Jinyuan Fang, Yanwen Peng, Xi~Zhang, Yingxu Wang, Xinhao Yi, Guibin Zhang, Yi~Xu, Bin Wu, Siwei Liu, Zihao Li, Zhaochun Ren, Nikos Aletras, Xi~Wang, Han Zhou, and Zaiqiao Meng. 2025.
\newblock \href {https://arxiv.org/abs/2508.07407} {A comprehensive survey of self-evolving ai agents: A new paradigm bridging foundation models and lifelong agentic systems}.
\newblock \emph{Preprint}, arXiv:2508.07407.

\bibitem[{Gill et~al.(2025)Gill, Ravichander, and Marasovic}]{gill-etal-2025-lost}
Alexander Gill, Abhilasha Ravichander, and Ana Marasovic. 2025.
\newblock \href {https://doi.org/10.18653/v1/2025.findings-emnlp.526} {What has been lost with synthetic evaluation?}
\newblock In \emph{Findings of the Association for Computational Linguistics: EMNLP 2025}, pages 9902--9945, Suzhou, China. Association for Computational Linguistics.

\bibitem[{Gong et~al.(2025)Gong, Mai, and Huang}]{gong2025mhierragmultimodalragvisualrich}
Ziyu Gong, Chengcheng Mai, and Yihua Huang. 2025.
\newblock \href {https://arxiv.org/abs/2508.00579} {Mhier-rag: Multi-modal rag for visual-rich document question-answering via hierarchical and multi-granularity reasoning}.
\newblock \emph{Preprint}, arXiv:2508.00579.

\bibitem[{Hofman et~al.(2026)Hofman, Brokman, Rachmil, Bose, Pahuja, Shimizu, Starostina, Marchisio, Goldfarb-Tarrant, and Vainshtein}]{hofman-etal-2026-maps}
Omer Hofman, Jonathan Brokman, Oren Rachmil, Shamik Bose, Vikas Pahuja, Toshiya Shimizu, Trisha Starostina, Kelly Marchisio, Seraphina Goldfarb-Tarrant, and Roman Vainshtein. 2026.
\newblock \href {https://doi.org/10.18653/v1/2026.findings-eacl.42} {{MAPS}: A multilingual benchmark for agent performance and security}.
\newblock In \emph{Findings of the {A}ssociation for {C}omputational {L}inguistics: {EACL} 2026}, pages 821--845, Rabat, Morocco. Association for Computational Linguistics.

\bibitem[{Huang et~al.(2024)Huang, Zhong, Lu, Zhu, Gao, Liu, Hou, Zeng, Wang, Shang, Jiang, Xu, and Liu}]{huang-etal-2024-planning-creation}
Shijue Huang, Wanjun Zhong, Jianqiao Lu, Qi~Zhu, Jiahui Gao, Weiwen Liu, Yutai Hou, Xingshan Zeng, Yasheng Wang, Lifeng Shang, Xin Jiang, Ruifeng Xu, and Qun Liu. 2024.
\newblock \href {https://doi.org/10.18653/v1/2024.findings-acl.259} {Planning, creation, usage: Benchmarking {LLM}s for comprehensive tool utilization in real-world complex scenarios}.
\newblock In \emph{Findings of the Association for Computational Linguistics: ACL 2024}, pages 4363--4400, Bangkok, Thailand. Association for Computational Linguistics.

\bibitem[{Kim et~al.(2026)Kim, Heo, Seo, Yeo, and Lee}]{kim2026agenticshopbenchmarkingagenticproduct}
Sunghwan Kim, Ryang Heo, Yongsik Seo, Jinyoung Yeo, and Dongha Lee. 2026.
\newblock \href {https://arxiv.org/abs/2602.12315} {Agenticshop: Benchmarking agentic product curation for personalized web shopping}.
\newblock \emph{Preprint}, arXiv:2602.12315.

\bibitem[{Koh et~al.(2024)Koh, Lo, Jang, Duvvur, Lim, Huang, Neubig, Zhou, Salakhutdinov, and Fried}]{koh-etal-2024-visualwebarena}
Jing~Yu Koh, Robert Lo, Lawrence Jang, Vikram Duvvur, Ming Lim, Po-Yu Huang, Graham Neubig, Shuyan Zhou, Russ Salakhutdinov, and Daniel Fried. 2024.
\newblock \href {https://doi.org/10.18653/v1/2024.acl-long.50} {{V}isual{W}eb{A}rena: Evaluating multimodal agents on realistic visual web tasks}.
\newblock In \emph{Proceedings of the 62nd Annual Meeting of the Association for Computational Linguistics (Volume 1: Long Papers)}, pages 881--905, Bangkok, Thailand. Association for Computational Linguistics.

\bibitem[{Lin et~al.(2025)Lin, Sun, Welleck, and Yang}]{lin2025leanstarlearninginterleavethinking}
Haohan Lin, Zhiqing Sun, Sean Welleck, and Yiming Yang. 2025.
\newblock \href {https://arxiv.org/abs/2407.10040} {Lean-star: Learning to interleave thinking and proving}.
\newblock \emph{Preprint}, arXiv:2407.10040.

\bibitem[{Liu et~al.(2025)Liu, Yu, Zhang, Xu, Lei, Lai, Gu, Ding, Men, Yang, Zhang, Deng, Zeng, Du, Zhang, Shen, Zhang, Su, Sun, Huang, Dong, and Tang}]{liu2025agentbenchevaluatingllmsagents}
Xiao Liu, Hao Yu, Hanchen Zhang, Yifan Xu, Xuanyu Lei, Hanyu Lai, Yu~Gu, Hangliang Ding, Kaiwen Men, Kejuan Yang, Shudan Zhang, Xiang Deng, Aohan Zeng, Zhengxiao Du, Chenhui Zhang, Sheng Shen, Tianjun Zhang, Yu~Su, Huan Sun, and 3 others. 2025.
\newblock \href {https://arxiv.org/abs/2308.03688} {Agentbench: Evaluating llms as agents}.
\newblock \emph{Preprint}, arXiv:2308.03688.

\bibitem[{Ma et~al.(2024)Ma, Zhang, Zhu, Yang, Yang, Jin, Lan, Kong, and He}]{ma2024agentboardanalyticalevaluationboard}
Chang Ma, Junlei Zhang, Zhihao Zhu, Cheng Yang, Yujiu Yang, Yaohui Jin, Zhenzhong Lan, Lingpeng Kong, and Junxian He. 2024.
\newblock \href {https://arxiv.org/abs/2401.13178} {Agentboard: An analytical evaluation board of multi-turn llm agents}.
\newblock \emph{Preprint}, arXiv:2401.13178.

\bibitem[{Mathew et~al.(2021)Mathew, Karatzas, and Jawahar}]{mathew2021docvqadatasetvqadocument}
Minesh Mathew, Dimosthenis Karatzas, and C.~V. Jawahar. 2021.
\newblock \href {https://arxiv.org/abs/2007.00398} {Docvqa: A dataset for vqa on document images}.
\newblock \emph{Preprint}, arXiv:2007.00398.

\bibitem[{Wang et~al.(2025{\natexlab{a}})Wang, Tao, Chen, Hu, and Qin}]{wang-etal-2025-x}
Peng Wang, Ruihan Tao, Qiguang Chen, Mengkang Hu, and Libo Qin. 2025{\natexlab{a}}.
\newblock \href {https://doi.org/10.18653/v1/2025.findings-acl.988} {{X}-{W}eb{A}gent{B}ench: A multilingual interactive web benchmark for evaluating global agentic system}.
\newblock In \emph{Findings of the Association for Computational Linguistics: ACL 2025}, pages 19320--19335, Vienna, Austria. Association for Computational Linguistics.

\bibitem[{Wang et~al.(2025{\natexlab{b}})Wang, Han, Diaz, Xu, Rühle, and Rajmohan}]{wang2025odysseybenchevaluatingllmagents}
Weixuan Wang, Dongge Han, Daniel~Madrigal Diaz, Jin Xu, Victor Rühle, and Saravan Rajmohan. 2025{\natexlab{b}}.
\newblock \href {https://arxiv.org/abs/2508.09124} {Odysseybench: Evaluating llm agents on long-horizon complex office application workflows}.
\newblock \emph{Preprint}, arXiv:2508.09124.

\bibitem[{Wang et~al.(2023)Wang, Kordi, Mishra, Liu, Smith, Khashabi, and Hajishirzi}]{wang2023selfinstructaligninglanguagemodels}
Yizhong Wang, Yeganeh Kordi, Swaroop Mishra, Alisa Liu, Noah~A. Smith, Daniel Khashabi, and Hannaneh Hajishirzi. 2023.
\newblock \href {https://arxiv.org/abs/2212.10560} {Self-instruct: Aligning language models with self-generated instructions}.
\newblock \emph{Preprint}, arXiv:2212.10560.

\bibitem[{Wang et~al.(2024)Wang, Cui, Zhong, Zhang, Yin, Lin, and Shang}]{wang2024officebenchbenchmarkinglanguageagents}
Zilong Wang, Yuedong Cui, Li~Zhong, Zimin Zhang, Da~Yin, Bill~Yuchen Lin, and Jingbo Shang. 2024.
\newblock \href {https://arxiv.org/abs/2407.19056} {Officebench: Benchmarking language agents across multiple applications for office automation}.
\newblock \emph{Preprint}, arXiv:2407.19056.

\bibitem[{Xie et~al.(2024)Xie, Zhang, Chen, Li, Zhao, Cao, Hua, Cheng, Shin, Lei, Liu, Xu, Zhou, Savarese, Xiong, Zhong, and Yu}]{xie2024osworldbenchmarkingmultimodalagents}
Tianbao Xie, Danyang Zhang, Jixuan Chen, Xiaochuan Li, Siheng Zhao, Ruisheng Cao, Toh~Jing Hua, Zhoujun Cheng, Dongchan Shin, Fangyu Lei, Yitao Liu, Yiheng Xu, Shuyan Zhou, Silvio Savarese, Caiming Xiong, Victor Zhong, and Tao Yu. 2024.
\newblock \href {https://arxiv.org/abs/2404.07972} {Osworld: Benchmarking multimodal agents for open-ended tasks in real computer environments}.
\newblock \emph{Preprint}, arXiv:2404.07972.

\bibitem[{Xu et~al.(2025)Xu, Sun, Zheng, Geng, Zhao, Feng, Tao, Lin, and Jiang}]{xu2025wizardlmempoweringlargepretrained}
Can Xu, Qingfeng Sun, Kai Zheng, Xiubo Geng, Pu~Zhao, Jiazhan Feng, Chongyang Tao, Qingwei Lin, and Daxin Jiang. 2025.
\newblock \href {https://arxiv.org/abs/2304.12244} {Wizardlm: Empowering large pre-trained language models to follow complex instructions}.
\newblock \emph{Preprint}, arXiv:2304.12244.

\bibitem[{Yang et~al.(2023)Yang, Prabhakar, Narasimhan, and Yao}]{yang2023intercodestandardizingbenchmarkinginteractive}
John Yang, Akshara Prabhakar, Karthik Narasimhan, and Shunyu Yao. 2023.
\newblock \href {https://arxiv.org/abs/2306.14898} {Intercode: Standardizing and benchmarking interactive coding with execution feedback}.
\newblock \emph{Preprint}, arXiv:2306.14898.

\bibitem[{Yao et~al.(2023)Yao, Chen, Yang, and Narasimhan}]{yao2023webshopscalablerealworldweb}
Shunyu Yao, Howard Chen, John Yang, and Karthik Narasimhan. 2023.
\newblock \href {https://arxiv.org/abs/2207.01206} {Webshop: Towards scalable real-world web interaction with grounded language agents}.
\newblock \emph{Preprint}, arXiv:2207.01206.

\bibitem[{Yue et~al.(2026)Yue, Bhandari, Ko, Patel, Lin, Zhou, Gao, Chen, and Pan}]{yue2026statictemplatesdynamicruntime}
Ling Yue, Kushal~Raj Bhandari, Ching-Yun Ko, Dhaval Patel, Shuxin Lin, Nianjun Zhou, Jianxi Gao, Pin-Yu Chen, and Shaowu Pan. 2026.
\newblock \href {https://arxiv.org/abs/2603.22386} {From static templates to dynamic runtime graphs: A survey of workflow optimization for llm agents}.
\newblock \emph{Preprint}, arXiv:2603.22386.

\bibitem[{Zhang et~al.(2026)Zhang, Acosta, Carlson, Bron, Doulcet, Ospina, and Suo}]{zhang2026parsebenchdocumentparsingbenchmark}
Boyang Zhang, Sebastián~G. Acosta, Preston Carlson, Sacha Bron, Pierre-Loïc Doulcet, Daniel~B. Ospina, and Simon Suo. 2026.
\newblock \href {https://arxiv.org/abs/2604.08538} {Parsebench: A document parsing benchmark for ai agents}.
\newblock \emph{Preprint}, arXiv:2604.08538.

\bibitem[{Zhou et~al.(2024)Zhou, Xu, Zhu, Zhou, Lo, Sridhar, Cheng, Ou, Bisk, Fried, Alon, and Neubig}]{zhou2024webarenarealisticwebenvironment}
Shuyan Zhou, Frank~F. Xu, Hao Zhu, Xuhui Zhou, Robert Lo, Abishek Sridhar, Xianyi Cheng, Tianyue Ou, Yonatan Bisk, Daniel Fried, Uri Alon, and Graham Neubig. 2024.
\newblock \href {https://arxiv.org/abs/2307.13854} {Webarena: A realistic web environment for building autonomous agents}.
\newblock \emph{Preprint}, arXiv:2307.13854.

\end{thebibliography}

\appendix
\clearpage

\section{Model Versions}
\label{app:model_versions}

\begin{table}[h]
\centering
\tiny
\setlength{\tabcolsep}{3pt}
\renewcommand{\arraystretch}{1.12}
\begin{tabularx}{\columnwidth}{@{}p{2.45cm}X@{}}
\toprule
\textbf{Model} & \textbf{Checkpoint / version} \\
\midrule
Gemini-3.1-Pro & Gemini API (\texttt{gemini-3.1-pro}). \\
Gemini-3.5-Flash & Gemini API (\texttt{gemini-3.5-flash}). \\
GPT-5 & OpenAI Responses API (\texttt{gpt-5}). \\
GPT-4o & OpenAI API (\texttt{gpt-4o}). \\
Llama-3.3-70B & \texttt{meta-llama/Llama-3.3-70B-Instruct}. \\
Llama-3.1-8B & \texttt{meta-llama/Llama-3.1-8B-Instruct}. \\
EuroLLM-9B & \texttt{EuroLLM/EuroLLM-9B-Instruct}. \\
Qwen3-32B and 4B & \texttt{Qwen/Qwen3-32B and 4B}. \\
\bottomrule
\end{tabularx}
\caption{We evaluate proprietary models via official APIs and open-weight models locally on an architecture equipped with two A40GPUs(48GBVRAM).}
\label{tab:model_versions}
\end{table}

\section{Prompts and Execution Details}
\label{app:implementation}
\bench’s execution and evaluation take place through a series of steps for language, task and scenario where the agent receives task, apps, execution history, and action schema. 
For construction, it must return one action encoded in JSON. Responses that do not contain a JSON trigger a retry with an explicit reminder, and the action is counted as malformed when the retry fails. Malformed actions leave the sandbox unchanged, and the agent may recover on the next step. Files are reachable under \path{data}, which is rewritten to the real sandbox path before each command executes. 

\paragraph{Agent Message.}
At the beginning of the interactions, we define the context via a system message. It is defined once per task from the persona (context and user), and the introduction line of each available app.

\begin{tcolorbox}[colback=softgray,colframe=darkgray,boxrule=0.5pt, title=Agent Message]
\small
\textit{You are an AI assistant acting for the following user:}
\{persona\}\\
You can interact with an operating system and use apps to solve the task.\\
You must follow the instructions and use the JSON.\\
You can only generate one action at a time.\\
Your response \textbf{MUST} be a single JSON action and \textbf{NOTHING ELSE}: no explanation, no markdown. Output only the JSON.\\
Once you have read the information you need, take the action that modifies or creates the required file. Reading the same file twice is wasteful; prefer acting over inspecting.\\
To edit a Word or Excel file, use the dedicated word/excel app actions (for example word write\_to\_file).\\
When the task is complete, finish with \{'app': 'system', 'action': 'finish\_task'\}.\\
You can find files for your task in \texttt{/testbed/data}. If you don't know the filenames, please switch to shell app and call commands to list the directory.\\

\end{tcolorbox}

\paragraph{App-Selection Prompt.}
Before any app is selected, the agent is asked to switch to one of the available apps. 

\begin{tcolorbox}[colback=softgray,colframe=darkgray,boxrule=0.5pt,
title=App-Selection Prompt]
\small
\textit{\#Task: \{task\}\\
\#History: \{history\}\\
\#Available apps: \{available\_apps\}\\
\#Instruction:\\
~- choose an app from the available apps: \{'app': 'system', 'action': 'switch\_app', 'target\_app': [THE\_APP\_YOU\_CHOOSE]\}\\
\#Command:}
\end{tcolorbox}

\paragraph{Action-Selection Prompt.}
Once an app is active, the action space is restricted to that app's operations, augmented by app switching and task termination. The demonstration line of every available action is inlined in \{detailed\_instruction\}.

\begin{tcolorbox}[colback=softgray,colframe=darkgray,boxrule=0.5pt,
title=Action-Selection Prompt]
\small
\textit{\#Task: \{task\}\\
\#History: \{history\}\\
\#Current apps: \{current\_app\}\\
\#Instruction: Choose one action from the list as the next step.\\
\{detailed\_instruction\}\\
~- switch to another app among \{available\_apps\}: \{'app': 'system', 'action': 'switch\_app', 'target\_app': [THE\_APP\_YOU\_CHOOSE]\}\\
~- finish the task with your answer as None if the task is not a question: \{'app': 'system', 'action': 'finish\_task', 'answer': 'None'\}\\
~- finish the task with your answer if the task is a question: \{'app': 'system', 'action': 'finish\_task', 'answer': [ANSWER]\}\\
\#Command:}
\end{tcolorbox}

Each history entry is rendered as \texttt{- Step $i$: \{action\} -> [\{observation\}]}, the agent conditions every decision on the full sequence of previous actions and their observed outputs.

\section{LLM-as-a-Judge Evaluation}
\label{app:judge}

Two evaluation functions, \texttt{evaluate\_email} and \texttt{evaluate\_note}, are judged by a held-out model. First, they perform a case-insensitive keyword check on the extracted document text and fail directly if a required keyword is missing. The surviving text is then passed to the judge together with the task-specific criteria. The judge runs at temperature $0$ with an output cap of 10 tokens, and its verdict is read as a pass if the response contains \textsc{yes}.

\begin{tcolorbox}[colback=softgray,colframe=darkgray,boxrule=0.5pt,
title=Judge Prompt]
\small
\textit{[system] You are a strict evaluation judge.\\[2pt]
[user] You are an evaluator for a benchmark. You are given a document and a set of criteria. Decide whether the document fully satisfies ALL the criteria.\\
Document type: \{doc\_type\}\\
Criteria: \{criteria\}\\
Document content:\\
-----\\
\{document\_text\}\\
-----\\
Answer with a single word: YES if the document satisfies all the criteria, or NO otherwise. Do not output anything else.}
\end{tcolorbox}

When a task supplies no explicit criteria, the following defaults are used.

\begin{tcolorbox}[colback=softgray,colframe=darkgray,boxrule=0.5pt,
title=Default Judging Criteria]
\small
\textit{\textbf{\texttt{evaluate\_email}}: The text is a coherent, well-formed email written in natural language. It has an appropriate greeting and closing, a clear and relevant message body, and a consistent tone suitable for its purpose.\\[4pt]
\textbf{\texttt{evaluate\_note}}: The text is a coherent, well-formed note written in natural language. It clearly conveys the intended information in a concise and relevant manner.}
\end{tcolorbox}

\section{Deterministic Evaluation Functions}
\label{app:det_eval}

The five deterministic functions read the sandbox's final state and verify its exact properties. \texttt{evaluate\_contain} extracts the text of the target file according to its document type spreadsheets, Word documents, PDFs, plain text, calendars, and mailboxes) and then checks that every required keyword occurs as a case-insensitive substring; numeric keywords are compared after thousands separators are stripped. \texttt{evaluate\_not\_contain} is its negation. \texttt{evaluate\_file\_exist} checks that a required output file has been created. \texttt{evaluate\_excel\_cell\_value} verifies that each declared cell holds the expected value at the declared row and column. \texttt{evaluate\_calendar\_no\_overlap} parses the user's calendar, sorts the events by start time, and fails when any event ends after the next one begins.

\section{Execution Hyper-parameters}
\label{app:hyperparams}

\begin{table}[h]
\centering
\scriptsize
\setlength{\tabcolsep}{3pt}
\renewcommand{\arraystretch}{1.12}
\begin{tabularx}{\columnwidth}{@{}p{3.4cm}X@{}}
\toprule
\textbf{Parameter} & \textbf{Value} \\
\midrule
Iteration cap & 30 steps per task. \\
Actions per step & Exactly one JSON action object. \\
Agent temperature & [0,1], with 1024 max tokens. \\
Judge temperature & [0,1], with 10 max tokens. \\
Malformed-action retry & One retry with an explicit JSON reminder. \\
Stagnation detection & 5 identical consecutive actions trigger \texttt{got\_stuck}. \\
Sandbox & Fresh copy of the testbed per task. \\
Result writing & Atomic writes after every task. \\
Distractors & 15--20 per testbed. \\
\bottomrule
\end{tabularx}
\caption{Execution settings. We use the most deterministic temperature, based on the documentation for the model used.}
\label{tab:hyperparams}
\end{table}

Generally, trajectories terminate in one of three ways. The agent emits \texttt{finish\_task}, which counts as a clean termination. The agent repeats an identical action more times in a row, which is detected by a sliding window and converted into a \texttt{got\_stuck} signal. The agent reaches the iteration cap, which is recorded as an iteration-cap hit and reported in Table~\ref{tab:headline}.

\section{Benchmark Construction and Human Audit}
\label{app:audit}

\subsection{Human-Written Seeds}

For each language--locale setting, construction begins from a
human-written core of 20 culturally situated personas. Each persona specifies a role, location, language, and relevant working context.
Human contributors write and supervise four realistic tasks for each persona, yielding 80 human-written seed tasks per setting.

These seeds define the intended relationships between persona
characteristics, local context, task requirements, testbed artefacts, and evaluation targets. They also provide examples for the subsequent augmentation stage.

\subsection{Persona and Task Augmentation}

Starting from the human-written seed pool, the construction pipeline generates 30 additional personas for each language--locale setting.
Each augmented persona is associated with five or six candidate tasks, producing between 150 and 180 augmented task candidates per setting.

Together, the human-written and augmented pools contain 50 personas and between 230 and 260 candidate tasks for each setting. Candidate personas and tasks are generated directly within their target language and locale. 

\begin{table}[t]
\centering
\small
\setlength{\tabcolsep}{4pt}
\begin{tabular}{@{}lcc@{}}
\toprule
\textbf{Construction stage} &
\textbf{Per setting} &
\textbf{Total} \\
\midrule
Human-written personas & 20 & 160 \\
Human-written seed tasks & 80 & 640 \\
Augmented personas & 30 & 240 \\
Augmented task candidates & 150--180 & 1,200--1,440 \\
Candidate tasks before audit & 230--260 & 1,840--2,080 \\
\midrule
Retained personas & 50 & 400 \\
Retained tasks & 200 & 1,600 \\
\bottomrule
\end{tabular}
\caption{Construction and filtering of \bench across the eight
language--locale settings. Candidate-task totals vary because each
augmented persona is associated with five or six tasks.}
\label{tab:construction-stats}
\end{table}

\subsection{Human Audit Procedure}

The candidate tasks are reviewed by annotators with relevant language and cultural expertise before inclusion in the final benchmark. Annotators inspect the persona, task instruction, supplied artefacts, and expected outcome.

The audit examines whether each task is realistic for the corresponding persona, appropriate for its local context, and solvable from the provided files. Annotators additionally assess whether the type and amount of information contained in each artefact are plausible for the corresponding document format.
Free-text feedback is requested whenever a task is judged unrealistic, culturally inappropriate, insufficiently specified, or unsupported by the supplied artefacts. This feedback is used to revise task instructions, testbed contents, and evaluation criteria, or to remove the candidate task when the identified problem cannot be resolved.

The audit involved annotators, with 40 assigned to each language--locale setting. Each candidate task received one annotation, and 25\% of the pool was double-annotated. Disagreements were resolved via internal author discussion. Annotators were compensated at 12(GBP/h).  Annotators were born in and residents of the country corresponding to the target setting, were fluent in the relevant language, and had sufficient digital literacy to inspect documents, spreadsheets, CSV files, and PDFs. They inspected the persona, task instruction, artefacts, and expected outcome.

\paragraph{Annotation Questions}

For each task, annotators answer the following questions:

\begin{enumerate}[leftmargin=*,itemsep=2pt]
    \item Is this a realistic task that the persona is likely to perform under the described circumstances?

    \item Is the task realistic with respect to the places, laws, organisations, culture, and location-specific characteristics described in the scenario?

    \item What information contained in each supplied file is necessary to complete the task?

    \item Is the type of information realistic for this document format?

    \item Is the amount of information realistic for this document format?
\end{enumerate}

Negative judgements are accompanied by free-text feedback that describes the problem and the required revision.

\subsection{Filtering and Final Balancing}

The annotation results are used as a filtering and revision mechanism. Tasks with recoverable problems are revised and checked again, whereas invalid or culturally implausible instances are removed. Following human review, the benchmark is balanced to 200 tasks and 50 distinct personas for each of the eight language--locale settings.
The final benchmark therefore contains 1,600 tasks and 400 personas, with an average of four retained tasks per persona.

\section{Step Budget}
\label{app:step-budget}

The iteration cap bounds how much exploration an agent may perform, and a low \cts could reflect an insufficient budget instead of a genuine limitation. We rerun the benchmark with caps of 10, 20, 30, 40, and 50 steps and report \cts in Table~\ref{tab:stepbudget}.

\begin{table}[h]
\centering
\small
\setlength{\tabcolsep}{4pt}
\begin{tabular}{@{}lccccc@{}}
\toprule
\textbf{Model} & \textbf{10} & \textbf{20} & \textbf{30} & \textbf{40} & \textbf{50} \\
\midrule
Gemini-3.1-Pro & 38.4 & 47.1 & \textbf{49.2} & 49.6 & 49.5 \\
GPT-4o & 27.3 & 36.2 & \textbf{38.6} & 39.4 & 39.5 \\
Gemini-3.5-Flash & 21.6 & 31.5 & \textbf{34.6} & 35.8 & 36.0 \\
Llama-3.3-70B & 13.2 & 21.0 & \textbf{24.1} & 25.6 & 26.0 \\
Llama-3.1-8B & 6.1 & 10.4 & \textbf{12.5} & 13.4 & 13.7 \\
EuroLLM-9B & 5.2 & 8.9 & \textbf{10.8} & 11.6 & 11.9 \\
\bottomrule
\end{tabular}
\caption{\cts under increasing iteration caps. The column in bold is the cap used throughout the paper.}
\label{tab:stepbudget}
\end{table}

Performance increases between 10 and 30 steps, and between 30 and 50 steps; gains range from 0.3 to 1.9 \cts across the evaluated subset.
Agents that fail at 30 steps continue to fail at 50, since additional steps are spent on repeated inspections and unproductive edits instead of on recovery. The weaker models benefit slightly more from a larger budget, which is consistent with their higher iteration-cap hit rate in Table~\ref{tab:headline}, and even for them the extra budget converts into fewer than two additional points.

\section{Trajectory Length by Model}
\label{app:trajectory-length}

\begin{figure}[t]
    \centering
    \plotorbox{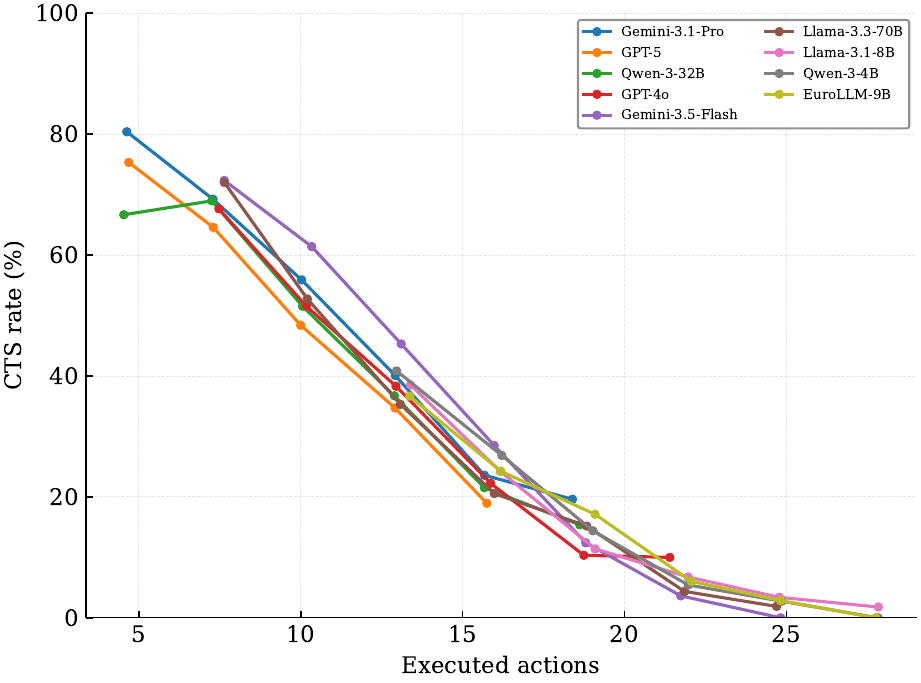}{\columnwidth}
    \caption{\cts rate by trajectory length for the evaluated models.}
    \label{fig:steps-all-models}
\end{figure}

Figure~\ref{fig:steps-all-models} examines the relationship between executed trajectory length and \cts for all nine evaluated models.
For each model, we pool the 1,600 task executions across the eight language-locale settings and group them into three-action-trajectory bins. Each marker reports the proportion of tasks achieving \cts within
the corresponding bin. Bins containing fewer than 20 task executions are omitted to reduce the influence of unstable estimates.

Across models, \cts generally decreases as trajectories become longer. The decline is particularly pronounced once execution exceeds the typical length of successful trajectories. Stronger models retain a higher success rate in the shorter bins, whereas weaker models more frequently produce long trajectories without reaching a valid and preserved final state.

\section{Judge Robustness}
\label{app:judge-robustness}

Two of the seven evaluation functions depend on an LLM judge, introducing a source of variance that the deterministic functions lack. We therefore rerun the judged evaluations with three-judge models and compare each against a human reference on 400 doubly annotated items.

\begin{table}[h]
\centering
\small
\setlength{\tabcolsep}{3pt}
\begin{tabular}{@{}lcccc@{}}
\toprule
\textbf{Judge} & \textbf{Email} & \textbf{Note} & \textbf{$\kappa$} & \textbf{Acc.} \\
\midrule
GPT-4o & 51.3 & 47.0 & 0.79 & 91.2 \\
Gemini-3.5 & 52.8 & 48.6 & 0.76 & 89.8 \\
Claude-Sonnet-4 & 50.4 & 46.2 & 0.81 & 92.4 \\
\midrule
Majority vote & 51.5 & 47.3 & 0.84 & 93.6 \\
Human reference & 49.7 & 45.1 & --- & --- \\
\bottomrule
\end{tabular}
\caption{Judge robustness on the two judged functions, averaged over models and languages. Email and Note report pass rate (\%). $\kappa$ is Cohen's kappa against the human reference, and Acc.\ is raw agreement (\%).}
\label{tab:judge}
\end{table}

The three judges agree with one another on 94.1\% of items, so the choice of judge shifts the reported pass rate by at most 2.4 points on either function. Agreement with the human reference is substantial throughout, with Cohen's kappa between 0.76 and 0.81, and a majority vote raises it to 0.84. All three judges are slightly more permissive than the human annotators, as expected, given that the default criteria reward well-formedness and do not penalise minor factual drift. The ranking of the valuated agents is identical under all three judges and under the human reference, so the conclusions of \S\ref{sec:results} do not depend on the judge.

\section{Recurring Error Patterns}
\label{app:error-patterns}

We inspect 400 failed trajectories, stratified by model and language, and group them into six recurring patterns. Table~\ref{tab:errorpatterns} reports the share of failures attributable to each pattern split by language.

\begin{table}[h]
\centering
\small
\setlength{\tabcolsep}{3pt}
\begin{tabular}{@{}lccc@{}}
\toprule
\textbf{Pattern} & \textbf{All} & \textbf{EN} & \textbf{non-EN} \\
\midrule
Distractor capture & 24.6 & 27.1 & 22.8 \\
Redundant inspection & 19.1 & 21.4 & 17.5 \\
Shell over-reach & 17.8 & 19.6 & 16.5 \\
Locale mismatch & 15.2 & 6.8 & \textbf{21.4} \\
Premature termination & 13.7 & 14.9 & 12.7 \\
Schema violation & 9.6 & 10.2 & 9.1 \\
\bottomrule
\end{tabular}
\caption{Recurring error patterns over manually inspected failed trajectories. Shares are computed within each column and sum to 100\%.}
\label{tab:errorpatterns}
\end{table}

We define the following patterns. \emph{Distractor capture} occurs when the agent edits a file whose name resembles the target, and a task asking for \texttt{budget\_2024.xlsx} leaves \texttt{budget\_2023.xlsx} modified as well. \emph{Redundant inspection} occurs when the agent reads the same artefact repeatedly without acting, which eventually triggers the stagnation signal. \emph{Shell over-reach} occurs when the agent issues a shell command where a dedicated app action exists, and a recursive copy or a wildcard removal touches several non-target files at once. \emph{Localisation mismatch} occurs when the agent writes values in the conventions of the wrong setting, such as an American date order in an Italian task or a decimal point where a decimal comma is expected. \emph{Premature termination} occurs when the agent emits \texttt{finish\_task} after satisfying only the first of several evaluation functions. \emph{Schema violation} occurs when the agent emits prose, a code fence, or an action with a missing argument.

We can report two main observations. First, distractor capture and shell over-reach together account for 42.4\% of failures, and both are preservation failures by construction, which explains the width of the gap between pass rate and \cts. Second, locale mismatch is the only pattern whose share varies substantially by language, rising from 6.8\% in English settings to 21.4\% in non-English settings and peaking at 26.9\% in Chinese. This pattern is therefore the principal mechanism underlying the language gradient in \S\ref{sec:results}, and it is a grounding failure, since the agents parse the instruction correctly and then emit values under the conventions of the wrong locale.

\begin{figure}[h]
    \centering
    \plotorbox{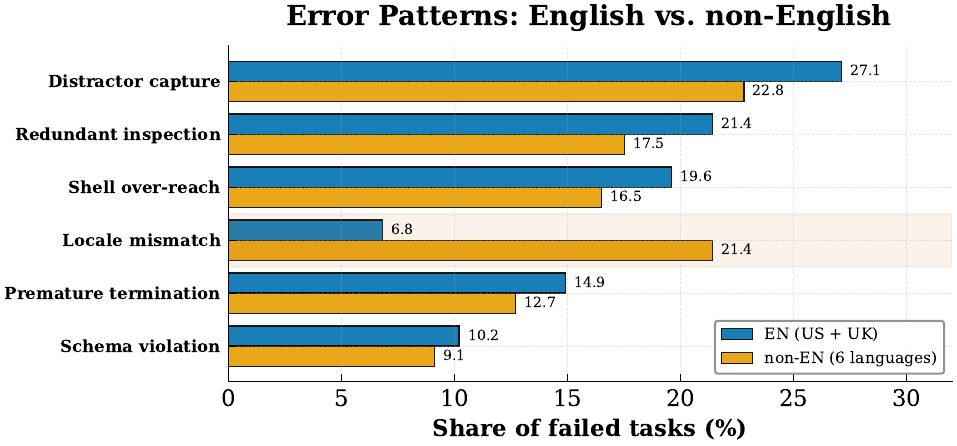}{\columnwidth}
    \caption{Failures by error pattern in the English and non-English settings.}
    \label{fig:errorlang}
\end{figure}

\section{Benchmark Composition}
\label{app:composition}

\paragraph{Task complexity.}
We measure the complexity of a task as the number of substantive actions in its reference solution, excluding app switches and the final termination action. A two-step task typically involves reading one artefact and applying one edit, whereas a five-step task requires the agent to gather information from several files before producing the required output. Figure~\ref{fig:complexity} reports the distribution. Tasks span two to five steps, with three-step tasks forming the largest group (32.5\%) and five-step tasks the smallest (16.5\%), for a mean reference length of 3.3 steps. The observed trajectories in \S\ref{sec:results} are longer than the reference solutions, with about 7/8 steps for the strongest model on solved tasks, which indicates that even successful agents spend most of their budget on exploration.

\begin{figure}[h]
    \centering
    \plotorbox{figures/task_complexity.pdf}{0.86\columnwidth}
    \caption{Distribution of task complexity, measured as the number of actions in the reference solution.}
    \label{fig:complexity}
\end{figure}

\begin{figure}[h]
    \centering
    \plotorbox{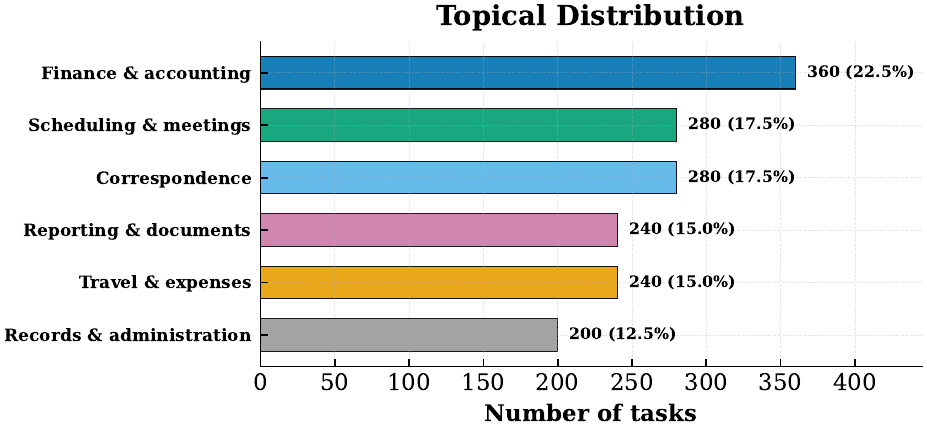}{\columnwidth}
    \caption{Topic distribution.}
    \label{fig:topics}
\end{figure}

\paragraph{Topics.}
Each task inherits a topic from the concerns of its Persona. Figure~\ref{fig:topics} groups the 1,600 tasks into six topic families. Finance and accounting tasks, which centre on spreadsheet manipulation, form the largest family, followed by scheduling and correspondence, which exercise the calendar and email tools. Reporting tasks combine reading across artefacts with document authoring; travel and expense tasks mix spreadsheets with correspondence; and records and administration tasks concern file organisation and retrieval. Every topic family appears in every language setting.

\paragraph{Evaluation targets.}
Table~\ref{tab:eval_usage} reports how often each evaluation function is attached to a task. Content containment is the most frequent target, since most tasks require specific information to appear in a specific artefact, and it is often combined with its negation to verify that distractor content was not propagated. Spreadsheet cell checks and file existence cover the structured and generative tasks respectively, and the two judge-based functions account for slightly over a tenth of all attachments, which bounds the influence of the judge on the headline scores.

\begin{table}[h]
\centering
\scriptsize
\setlength{\tabcolsep}{3pt}
\renewcommand{\arraystretch}{1.12}
\begin{tabularx}{\columnwidth}{@{}p{3.0cm}cX@{}}
\toprule
\textbf{Function} & \textbf{Share} & \textbf{Typical target} \\
\midrule
\texttt{contain} & 41\% & Required content in the target artefact \\
\texttt{excel\_cell\_value} & 18\% & Structured numeric edits \\
\texttt{file\_exist} & 14\% & Creation of a required output file \\
\texttt{not\_contain} & 9\% & Distractor content not propagated \\
\texttt{calendar\_no\_overlap} & 7\% & Consistent scheduling \\
\texttt{evaluate\_email} & 6\% & Communicative adequacy of emails \\
\texttt{evaluate\_note} & 5\% & Communicative adequacy of notes \\
\bottomrule
\end{tabularx}
\caption{Share of evaluation functions.}
\label{tab:eval_usage}
\end{table}

\paragraph{Language balance.}
The benchmark is balanced at the aggregate level; each language--locale setting contains 200 tasks, with the same distribution of reference trajectory lengths. We additionally control the distributions of topic families, application types, and evaluation functions across settings.

\paragraph{Cross-Locale Comparability}
\label{app:cross-locale}

The task sets are independently constructed within each setting, without translations or item-level counterparts of tasks in the other settings. We instead balance their aggregate distributions with respect to the number of tasks, topic families, application types, evaluation functions, and reference trajectory lengths.
This construction preserves native language use and locale-specific conventions. They may reflect a combination of multilingual instruction following, localised document conventions, cultural grounding, and residual task variation, and should not be interpreted as isolated causal effects of language.

\newtcolorbox{wbexample}[1][]{
  colback=black!2, colframe=black!55, boxrule=0.5pt, arc=1pt, left=4pt, right=4pt, top=4pt, bottom=4pt, #1}

\section{Examples}
\label{app:examples}

We report three examples of \bench{}, from the EN-UK, FR, and IT. Each instance shows the persona, the native-language instruction, the evaluation configuration, and the trajectory executed in the sandbox. We report some inference from Gemini-3.1-Pro as the backbone model.

\paragraph{Successful execution.}
Figure~\ref{fig:example-en} reports an EN-UK instance evaluated by one deterministic function and one judge-based function. The agent inspects the delivery note, writes the required status into the target cell, and sets the note requested by the instruction, while the superseded copy of the ledger remains untouched. Both functions succeed, and the 17 pre-existing non-target artefacts remain identical, hence $\mathrm{pass}(t)\wedge\mathrm{preserve}(t)$ holds and the task receives \cts.
Figure~\ref{fig:example-fr} reports a French instance consisting of two applications and evaluated by a spreadsheet-cell check together with a containment check over the mailbox. The retained amount follows from the conventional mileage rate recorded in the testbed and is expressed with a decimal comma, which is the convention declared by the containment keyword; hence an agent that emits \texttt{42.56} fails the check even when the underlying arithmetic is correct.

\paragraph{Failed execution.}
Figure~\ref{fig:example-it} reports an Italian instance in which the task-specific evaluator succeeds whilst preservation fails. The testbed contains a superseded copy of the deadline register, and the agent writes the required status into that copy before locating the current one. The deterministic check inspects the target file alone and returns $\mathrm{pass}(t)=\mathrm{true}$, whereas \texttt{scadenze\_pratiche\_2025.xlsx} differs from its initial state, which gives $\mathrm{preserve}(t)=\mathrm{false}$ and $\mathrm{CTS}(t)=0$. This is the distractor-capture pattern of Appendix~\ref{app:error-patterns}, which accounts for 24.6\% of the manually inspected failures and remains invisible to correctness-only scoring. The lower part of the same figure reports a locale-mismatch variant, where the instruction additionally requires a reminder letter carrying the outstanding balance, and the agent recovers the correct figure whilst emitting it under the conventions of the wrong locale, which causes the containment check to fail on a preserved workspace.

\begin{figure*}[t]
\centering
\begin{wbexample}
\footnotesize
\textbf{Setting} EN-UK \quad\textbf{Applications} shell, excel, word \quad
\textbf{Testbed} 18 artefacts, 16 distractors \quad
\textbf{Outcome} \cts $=1$

\medskip
\textbf{Persona.} Rhodri manages stock control for a community pharmacy in Cardiff. He reconciles wholesaler delivery notes against the branch ledger, records short deliveries for the responsible pharmacist, and keeps the controlled-drug paperwork ready for inspection.

\textbf{Task.} Reconcile the March wholesaler delivery note with the branch
stock ledger, record the short-delivered line in the ledger, and leave a note
for the responsible pharmacist naming the affected product.

\medskip
\textbf{Evaluation configuration.}
\begin{Verbatim}[fontsize=\scriptsize,baselinestretch=0.92,xleftmargin=6pt]
[{"function": "evaluate_excel_cell_value",
  "args": {"file": "./data/stock_ledger_cardiff.xlsx",
           "matches": [{"row": 4, "col": 3, "value": "Short delivery"}]}},
 {"function": "evaluate_note",
  "args": {"file": "./data/short_delivery_note.docx", "doc_type": "docx",
           "keywords": ["Amoxicillin 500mg", "short delivery"]}}]
\end{Verbatim}

\textbf{Executed trajectory.}
\begin{Verbatim}[fontsize=\scriptsize,baselinestretch=0.92,xleftmargin=6pt]
Step 0: {'app': 'system', 'action': 'switch_app', 'target_app': 'shell'}
     -> [Successfully switched to app: shell]
Step 1: {'app': 'shell', 'action': 'command', 'command': 'ls data'}
     -> [cd_destruction_log.txt  controlled_drugs_register.xlsx
         counter_stock_note.txt  credit_claims_2025.xlsx
         delivery_note_february_2026.txt  delivery_note_march_2026.txt  ...
         stock_ledger_cardiff.xlsx  stock_ledger_cardiff_2025.xlsx
         stock_take_february_2026.xlsx  wholesaler_terms_cambrian.txt]
Step 2: {'app': 'shell', 'action': 'command',
         'command': 'cat data/delivery_note_march_2026.txt'}
     -> [Cambrian Medical Supplies - delivery note 12 March 2026
         Account: Pontcanna Community Pharmacy, Cardiff
         Paracetamol 500mg caps: ordered 100, delivered 100
         Amoxicillin 500mg caps: ordered 60, delivered 40
         Ibuprofen 200mg tabs: ordered 50, delivered 50]
Step 3: {'app': 'system', 'action': 'switch_app', 'target_app': 'excel'}
     -> [Successfully switched to app: excel]
Step 4: {'app': 'excel', 'action': 'read_file',
         'file_path': './data/stock_ledger_cardiff.xlsx'}
     -> [The following is the table from the excel file:
         (1, 1): Product  (1, 2): Ordered  (1, 3): Delivery status
         (2, 1): Paracetamol 500mg caps  (2, 2): 100  (2, 3): Received in full
         (3, 1): Ibuprofen 200mg tabs  (3, 2): 50  (3, 3): Received in full
         (4, 1): Amoxicillin 500mg caps  (4, 2): 60  (4, 3): Awaiting check]
Step 5: {'app': 'excel', 'action': 'set_cell',
         'file_path': './data/stock_ledger_cardiff.xlsx',
         'row_idx': 4, 'column_idx': 3, 'text': 'Short delivery'}
     -> [Successfully write text to ./data/stock_ledger_cardiff.xlsx]
Step 6: {'app': 'system', 'action': 'switch_app', 'target_app': 'word'}
     -> [Successfully switched to app: word]
Step 7: {'app': 'word', 'action': 'create_new_file',
         'file_path': './data/short_delivery_note.docx'}
     -> [Successfully create new file ./data/short_delivery_note.docx]
Step 8: {'app': 'word', 'action': 'write_to_file',
         'file_path': './data/short_delivery_note.docx',
         'contents': 'Note for the responsible pharmacist. The Cambrian Medical
          Supplies delivery of 12 March 2026 is a short delivery. Amoxicillin
          500mg caps: 40 units received against 60 ordered. The ledger has been
          updated and a credit claim will be raised for the outstanding 20
          units. Rhodri'}
     -> [Successfully write contents to ./data/short_delivery_note.docx]
Step 9: {'app': 'system', 'action': 'finish_task', 'answer': 'None'}
     -> [Task finished]
\end{Verbatim}

\textbf{Outcome.} \texttt{evaluate\_excel\_cell\_value} matches
\texttt{(4, 3): Short delivery} and \texttt{evaluate\_note} passes both the
keyword filter and the judge. No non-target artefact is modified, hence
$\mathrm{CTS}(t)=1$.
\end{wbexample}
\caption{EN-UK instance solved under both conditions of CTS. The agent reads the
evidence artefact once, edits the target cell, and creates the required note
without touching the superseded copy of the ledger.}
\label{fig:example-en}
\end{figure*}

\begin{figure*}[t]
\centering
\begin{wbexample}
\footnotesize
\textbf{Setting} FR \quad\textbf{Applications} excel, email \quad
\textbf{Testbed} 19 artefacts, 16 distractors \quad
\textbf{Outcome} \cts $=1$

\medskip
\textbf{Persona.} Amélie est coordinatrice administrative dans une association
d'aide à domicile à Lyon. Elle contrôle les notes de frais des intervenants au
regard des indemnités kilométriques conventionnelles, tient à jour le suivi des
remboursements et transmet les dossiers validés à la trésorière pour la paie.

\textbf{Task.} Vérifie la note de frais de Karim au regard des indemnités
kilométriques conventionnelles, mets à jour le suivi des remboursements et
préviens la trésorière par courriel du montant retenu.
\emph{(Gloss: check the expense claim against the conventional mileage
allowance, update the reimbursement tracker, and inform the treasurer of the
retained amount by email.)}

\medskip
\textbf{Evaluation configuration.}
\begin{Verbatim}[fontsize=\scriptsize,baselinestretch=0.92,xleftmargin=6pt]
[{"function": "evaluate_excel_cell_value",
  "args": {"file": "./data/suivi_remboursements_lyon.xlsx",
           "matches": [{"row": 3, "col": 4, "value": "Validé"}]}},
 {"function": "evaluate_contain",
  "args": {"doc_type": "email", "username": "amelie",
           "keywords": ["Karim Benali", "42,56"]}}]
\end{Verbatim}

\textbf{Executed trajectory.}
\begin{Verbatim}[fontsize=\scriptsize,baselinestretch=0.92,xleftmargin=6pt]
Step 0: {'app': 'system', 'action': 'switch_app', 'target_app': 'excel'}
     -> [Successfully switched to app: excel]
Step 1: {'app': 'excel', 'action': 'read_file',
         'file_path': './data/note_frais_karim.xlsx'}
     -> [The following is the table from the excel file:
         (1, 1): Intervenant  (1, 2): Mois  (1, 3): Kilomètres  (1, 4): Statut
         (2, 1): Karim Benali  (2, 2): mars 2026  (2, 3): 112
         (2, 4): À traiter]
Step 2: {'app': 'excel', 'action': 'read_file',
         'file_path': './data/indemnites_kilometriques_2026.xlsx'}
     -> [The following is the table from the excel file:
         (1, 1): Année  (1, 2): Véhicule  (1, 3): Tarif par km
         (1, 4): Plafond mensuel
         (2, 1): 2026  (2, 2): Automobile  (2, 3): 0,38  (2, 4): 250,00
         (3, 1): 2026  (3, 2): Deux-roues  (3, 3): 0,16  (3, 4): 250,00]
Step 3: {'app': 'excel', 'action': 'set_cell',
         'file_path': './data/suivi_remboursements_lyon.xlsx',
         'row_idx': 3, 'column_idx': 4, 'text': 'Validé'}
     -> [Successfully write text to ./data/suivi_remboursements_lyon.xlsx]
Step 4: {'app': 'system', 'action': 'switch_app', 'target_app': 'email'}
     -> [Successfully switched to app: email]
Step 5: {'app': 'email', 'action': 'send_email', 'sender': 'amelie',
         'recipient': 'claire',
         'subject': 'Note de frais mars 2026 - Karim Benali',
         'content': 'Bonjour Claire, la note de frais de Karim Benali pour mars
          2026 est validée. Le montant retenu est de 42,56 EUR, soit 112 km au
          tarif conventionnel de 0,38 EUR par km, sous le plafond mensuel de
          250,00 EUR. Le suivi est à jour, à intégrer à la paie de mars.
          Cordialement, Amélie'}
     -> [Successfully sent email to claire.]
Step 6: {'app': 'system', 'action': 'finish_task', 'answer': 'None'}
     -> [Task finished]
\end{Verbatim}

\textbf{Outcome.} The tracker cell matches \texttt{(3, 4): Validé} and the
mailbox of the persona contains both keywords, hence
$\mathrm{pass}(t)=\mathrm{true}$. The mileage table, the expense claim and the
remaining artefacts are unchanged, and the two \texttt{.eml} files written by
the send action are new artefacts, hence $\mathrm{CTS}(t)=1$.
\end{wbexample}
\caption{FR instance solved under both conditions of CTS. The retained amount is
derived from the conventional rate held in the testbed and written with a
decimal comma, which is the convention declared by the containment keyword.}
\label{fig:example-fr}
\end{figure*}

\begin{figure*}[t]
\centering
\begin{wbexample}
\footnotesize
\textbf{Setting} IT \quad\textbf{Applications} shell, excel \quad
\textbf{Testbed} 18 artefacts, 16 distractors \quad
\textbf{Outcome} \cts $=0$ with $\mathrm{pass}(t)=\mathrm{true}$

\medskip
\textbf{Persona.} Giulia è responsabile di segreteria in uno studio tecnico di
Bologna. Gestisce il registro delle pratiche edilizie, controlla i termini di
ultimazione dei lavori comunicati dallo Sportello Unico e prepara i solleciti.

\textbf{Task.} Controlla la pratica SCIA-2023-114 nel registro delle scadenze
dell'anno in corso e segna la pratica come scaduta.
\emph{(Gloss: check case SCIA-2023-114 in the current-year deadline register and
mark the case as expired.)}

\textbf{Evaluation configuration.}
\begin{Verbatim}[fontsize=\scriptsize,baselinestretch=0.92,xleftmargin=6pt]
[{"function": "evaluate_excel_cell_value",
  "args": {"file": "./data/scadenze_pratiche_2026.xlsx",
           "matches": [{"row": 5, "col": 3, "value": "Scaduta"}]}}]
\end{Verbatim}

\textbf{Executed trajectory.}
\begin{Verbatim}[fontsize=\scriptsize,baselinestretch=0.92,xleftmargin=6pt]
Step 0: {'app': 'system', 'action': 'switch_app', 'target_app': 'shell'}
     -> [Successfully switched to app: shell]
Step 1: {'app': 'shell', 'action': 'command', 'command': 'ls data'}
     -> [agenda_sopralluoghi.txt  computo_metrico_verdi.txt
         comunicazioni_comune.txt  corrispondenza_bianchi.txt  ...
         scadenze_pratiche_2025.xlsx  scadenze_pratiche_2026.xlsx  ...]
Step 2: {'app': 'system', 'action': 'switch_app', 'target_app': 'excel'}
     -> [Successfully switched to app: excel]
Step 3: {'app': 'excel', 'action': 'read_file',
         'file_path': './data/scadenze_pratiche_2025.xlsx'}
     -> [The following is the table from the excel file:
         (1, 1): Pratica  (1, 2): Committente  (1, 3): Stato  ...  (1, 5): Saldo
         ...  (5, 1): SCIA-2023-114  (5, 2): Bianchi S.r.l.  (5, 3): In corso
         (5, 4): 15/06/2026  (5, 5): 1.250,50]
Step 4: {'app': 'excel', 'action': 'set_cell',
         'file_path': './data/scadenze_pratiche_2025.xlsx',
         'row_idx': 5, 'column_idx': 3, 'text': 'Scaduta'}  <-- collateral edit
     -> [Successfully write text to ./data/scadenze_pratiche_2025.xlsx]
Step 5: {'app': 'excel', 'action': 'read_file',
         'file_path': './data/scadenze_pratiche_2026.xlsx'}
     -> [The following is the table from the excel file:
         (1, 1): Pratica  (1, 2): Committente  (1, 3): Stato  ...  (1, 5): Saldo
         ...  (5, 1): SCIA-2023-114  (5, 2): Bianchi S.r.l.  (5, 3): In corso
         (5, 4): 15/06/2026  (5, 5): 1.250,50]
Step 6: {'app': 'excel', 'action': 'set_cell',
         'file_path': './data/scadenze_pratiche_2026.xlsx',
         'row_idx': 5, 'column_idx': 3, 'text': 'Scaduta'}
     -> [Successfully write text to ./data/scadenze_pratiche_2026.xlsx]
Step 7: {'app': 'system', 'action': 'finish_task', 'answer': 'None'}
     -> [Task finished]
\end{Verbatim}

\textbf{Outcome.} The evaluator matches \texttt{(5, 3): Scaduta} in the target
register, hence $\mathrm{pass}(t)=\mathrm{true}$, whereas one of the 17
non-target artefacts differs from its initial state, which gives
$\mathrm{preserve}(t)=\mathrm{false}$ and $\mathrm{CTS}(t)=0$.

\medskip
\textbf{Locale-mismatch variant.} The instruction additionally requires the
reminder letter, and the configuration is extended with
\texttt{evaluate\_contain} on \texttt{./data/sollecito\_bianchi.docx} for the
keywords \texttt{"SCIA-2023-114"} and \texttt{"1.250,50"}.
\begin{Verbatim}[fontsize=\scriptsize,baselinestretch=0.92,xleftmargin=6pt]
Step 5: {'app': 'word', 'action': 'write_to_file',
         'file_path': './data/sollecito_bianchi.docx',
         'contents': 'Spett.le Bianchi S.r.l., il termine di ultimazione dei
          lavori della pratica SCIA-2023-114 e decorso il 15/06/2026. Risulta
          un saldo oneri di euro 1250.50 da versare entro trenta giorni.'}
     -> [Successfully write contents to ./data/sollecito_bianchi.docx]
\end{Verbatim}
The agent recovers the outstanding balance and writes it under conventions that
the setting does not use, which fails the containment check on an otherwise
preserved workspace and yields $\mathrm{pass}(t)=\mathrm{false}$ with
$\mathrm{CTS}(t)=0$.
\end{wbexample}
\caption{IT instance that satisfies the task-specific evaluator whilst violating
the preservation constraint. The collateral edit at Step~4 is the
distractor-capture pattern of Appendix~\ref{app:error-patterns}, and the variant
below reproduces the locale-mismatch pattern underlying the language.}
\label{fig:example-it}
\end{figure*}

\end{document}